\documentclass[10pt,twocolumn,letterpaper]{article}

\usepackage[T1]{fontenc}
\usepackage[utf8]{inputenc}
\usepackage{amsmath}
\usepackage{newtxtext,newtxmath}
\usepackage[letterpaper,textwidth=6.875in,textheight=8.875in,columnsep=0.3125in,top=1in,headsep=0.2in,footskip=0.4in]{geometry}
\usepackage{amsmath}
\usepackage[dvipsnames]{xcolor}
\usepackage{graphicx}
\usepackage{booktabs}
\usepackage{multirow}
\usepackage{xspace}
\usepackage[shortlabels,inline]{enumitem}
\usepackage[format=plain,labelformat=simple,labelsep=period,font=small,compatibility=false]{caption}
\usepackage{microtype}
\usepackage[ruled,vlined,linesnumbered]{algorithm2e}
\usepackage{placeins}
\usepackage[numbers,sort&compress]{natbib}
\usepackage[hyphens]{url}
\definecolor{linkblue}{rgb}{0.21,0.49,0.74}
\usepackage[pagebackref,breaklinks,colorlinks,allcolors=linkblue]{hyperref}
\usepackage[capitalize]{cleveref}
\crefname{section}{Sec.}{Secs.}
\Crefname{section}{Section}{Sections}
\crefname{appendix}{App.}{Apps.}
\Crefname{appendix}{Appendix}{Appendices}
\crefname{table}{Tab.}{Tabs.}
\Crefname{table}{Table}{Tables}
\crefname{algocf}{Alg.}{Algs.}
\Crefname{algocf}{Algorithm}{Algorithms}

\setlist[itemize]{noitemsep,leftmargin=*,topsep=0em}
\setlist[enumerate]{noitemsep,leftmargin=*,topsep=0em}
\renewcommand{\paragraph}[1]{\vspace{0.3em}\noindent\textbf{#1}}

\makeatletter
\DeclareRobustCommand\onedot{\futurelet\@let@token\@onedot}
\def\@onedot{\ifx\@let@token.\else.\null\fi\xspace}
\makeatother
\def\eg{\emph{e.g}\onedot} 
\def\ie{\emph{i.e}\onedot} 
\def\etal{\emph{et al}\onedot}

\newcommand{\method}{\texorpdfstring{\ensuremath{\nabla}Mimic}{NablaMimic}\xspace}
\newcommand{\engine}{NablaEngine\xspace}

\newcommand{\notrun}{n/r}
\newcommand{\notreached}{never}
\newcommand{\napo}{\texorpdfstring{\ensuremath{\nabla}PO}{NablaPO}\xspace}
\newcommand{\realClipCount}{5}
\newcommand{\realTrialCount}{7}
\newcommand{\realTotalFrames}{6,762}
\newcommand{\realTotalSec}{135.3}
\newcommand{\realJointRmsWeighted}{3.88}
\newcommand{\realJointRmsMin}{3.19}
\newcommand{\realJointRmsMax}{4.62}
\newcommand{\realTiltWeighted}{2.77}

\newcommand{\realTiltPninetyfiveMin}{3.7}
\newcommand{\realTiltPninetyfiveMax}{6.9}
\newcommand{\realTiltMaxClip}{12.8}

\newcommand{\realKPosWeighted}{0.994}

\newcommand{\realKOriWeighted}{0.949}

\newcommand{\realKRootWeighted}{0.982}
\newcommand{\realKAngWeighted}{0.967}
\newcommand{\realCtrlDtMed}{20.0}
\newcommand{\realCtrlDtMax}{21.0}
\newcommand{\realStateAgeMed}{0.5--0.6}
\newcommand{\realStateAgePninetyfivePlus}{1.8}
\newcommand{\realTauPninetyfiveMax}{0.47}
\newcommand{\realSatCtrlMax}{0.0}
\newcommand{\projecturl}{\url{https://sailab.ai/nablapo/}}

\hypersetup{
  pdftitle={NablaPO: Ultra-Efficient Policy Optimization Through Environment Gradients},
  pdfauthor={Weyl Lu, Gao Zhu, Weijie Lai, Zaishuo Xia, Yubei Chen},
}

\title{\Large\bfseries $\nabla$PO: Ultra-Efficient Policy Optimization Through Environment Gradients}
\author{Weyl Lu$^{1}$ \qquad Gao Zhu$^{1}$ \qquad Weijie Lai$^{1}$ \qquad Zaishuo Xia$^{1}$ \qquad Yubei Chen$^{1}$ \\[3pt]
\normalsize $^{1}$UC Davis \quad\textbullet\quad Project page: \projecturl}
\date{\vspace{-2.5em}}

\begin{document}
\twocolumn[{%
\renewcommand\twocolumn[1][]{#1}%
\maketitle
\begin{center}
  \captionsetup{type=figure}
  \vspace{-0.5em}
  \includegraphics[width=0.86\textwidth]{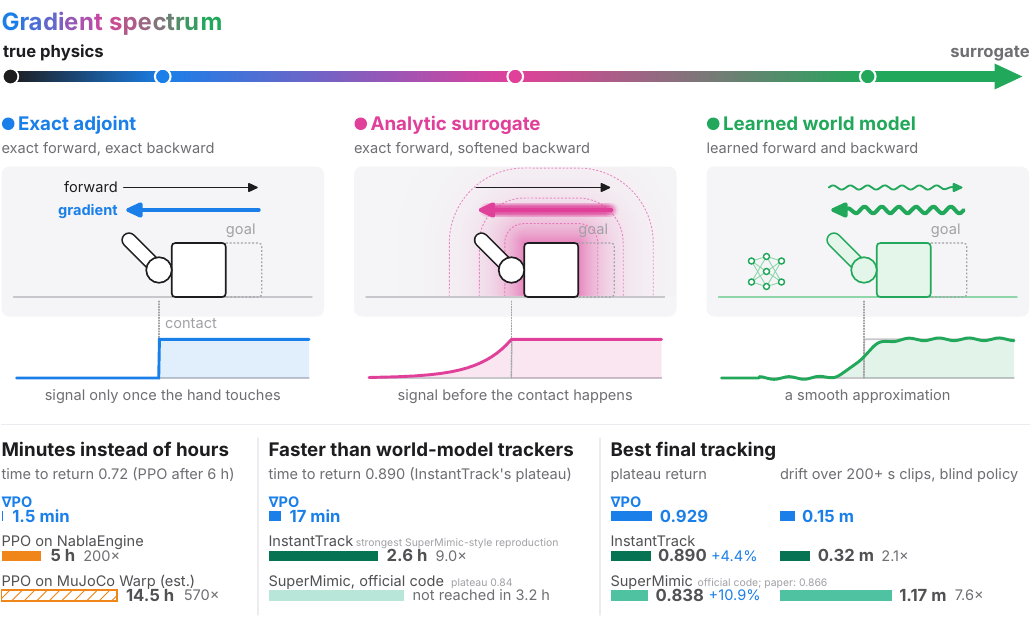}
  \captionof{figure}{\textbf{\napo turns multi-hour humanoid learning into minutes by backpropagating returns through environment gradients.} Top: three gradient sources for a hand pushing a box to a goal---\emph{exact adjoint} (simulator derivative, active upon contact), \emph{analytic surrogate} (exact forward rollout with a soft backward distance field for pre-contact signal), and \emph{learned world model} (learned forward and backward passes), with schematic action-gradient profiles below. Bottom: 40-clip LAFAN1 humanoid tracking with one policy on one GPU. Left: training time to the plateau of a 6\,h PPO run ($200\times$ faster; MuJoCo Warp scales \engine's time by their $2.9\times$ step-time ratio). Middle: time to InstantTrack's plateau ($9\times$ faster; SuperMimic official code does not reach it in 3.2\,h). Right: plateau return (+4.4\% over InstantTrack, +10.9\% over SuperMimic official code; its paper states 0.866) and mean root drift of blind policies on the 30 clips $>200$\,s ($2.1\times$ and $7.6\times$ lower drift).}
  \vspace{0.5em}
  \label{fig:teaser}
\end{center}%
}]
\begin{figure*}[t]
\centering
\includegraphics[width=0.90\linewidth]{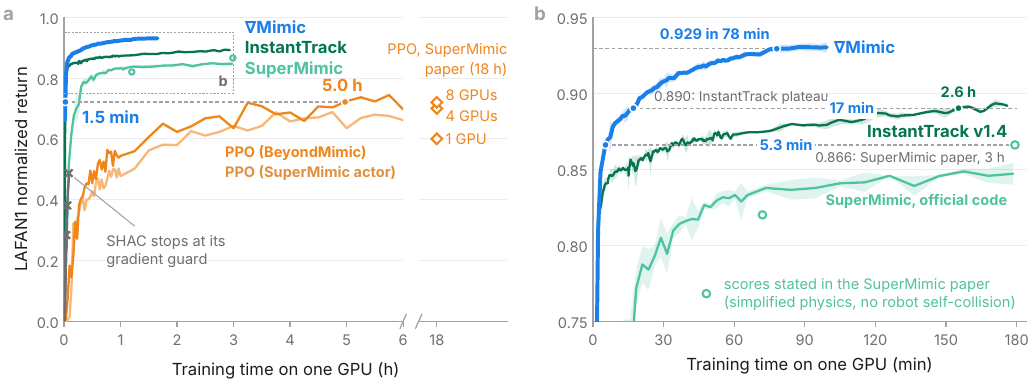}
\caption{\textbf{\method reaches the plateau return of a 6-hour PPO run $200\times$ faster (1.5\,min vs.\ 5.0\,h) and surpasses world-model trackers in both speed and asymptotic return on 40-clip LAFAN1 tracking.} (a) Normalized return vs.\ training time on one RTX PRO 6000 GPU for \method (mean $\pm$ s.d., 5 seeds), InstantTrack v1.4~\cite{dao2026instanttrack} (3 seeds, identical physics), SuperMimic~\cite{liao2026supermimic} official code (3 seeds), PPO in two configurations (1 seed each), and SHAC (3 seeds; crosses mark where its gradient guard halts training). Diamonds after the 6\,h axis break show PPO returns reported by SuperMimic after 18\,h on 1, 4, and 8 GPUs. (b) First 180 minutes (dashed box in (a)). Hollow circles denote scores reported in the SuperMimic paper (0.82 at 1.2\,h and 0.866 at 3\,h, trained without self-collision). Dashed horizontal links show \method reaching SuperMimic's 3\,h score in 5.3\,min ($34\times$ faster) and InstantTrack's 2.6\,h plateau in 17\,min ($9\times$ faster), before converging to 0.929 at 78\,min. Three further PPO seeds in the BeyondMimic configuration, trained for 48\,h (13.6--16.8\,h at the RTX PRO 6000 update rate), end at $0.782\pm0.005$ (\cref{tab:levels}).}
\label{fig:hero}
\end{figure*}

\begin{abstract}
Zeroth-order reinforcement learning methods such as PPO treat the simulator as a black box and require hours to days of compute for whole-body robot control, while existing first-order methods fail on contact-rich dynamics due to exploding gradients and slow differentiable simulators. We introduce \napo, a first-order policy optimization framework built on three pillars: \emph{per-sample gradient clipping} relative to the batch median, \emph{multi-step action-target regression} that forms an automatic action-space trust region, and \emph{ultrafast exact adjoints} in \engine---our GPU physics engine that computes exact MuJoCo contact adjoints \textbf{8--12$\times$ faster} and forward dynamics \textbf{5$\times$ faster} than MuJoCo Warp on current GPUs. Training a single policy across all 40 LAFAN1 humanoid motion clips on one GPU, \napo matches the plateau return of a 6-hour PPO run in \textbf{1.5 minutes} (\textbf{200$\times$ faster}), SuperMimic's reported 3-hour score in \textbf{5.3 minutes} (\textbf{34$\times$ faster}), and InstantTrack's 2.6-hour plateau in \textbf{17 minutes} (\textbf{9$\times$ faster}), while achieving a higher asymptotic return (\textbf{0.929 vs.\ 0.890}) with zero falls and \textbf{2.1--7.6$\times$ lower} root drift over 200\,s horizons; the tracker deploys directly from simulation to a physical Unitree G1 without real-world fine-tuning. Across 16 DeepMind Control and Brax benchmarks, \napo reaches 80\% of peak return faster than PPO on \textbf{14 of 16 tasks}, achieves the highest final return on \textbf{13 of 16}, and exceeds the final return of prior first-order methods on every task they were run on (SHAC: 16, AHAC: 15). The same learner acquires humanoid velocity tracking in \textbf{3 minutes}, 2\,kg box lifting in \textbf{10 minutes}, joint joystick-and-motion control in \textbf{11 minutes}, and 10-terrain traversal in \textbf{20 minutes}. By collapsing multi-hour training into minutes on a single GPU, \napo removes the compute bottleneck in whole-body robot learning and makes systematic scaling studies practical.
\end{abstract}
\section{Introduction}
\label{sec:intro}

Reinforcement learning for legged and humanoid robots relies predominantly on zeroth-order algorithms. PPO~\cite{schulman2017proximal} and related likelihood-ratio estimators treat the physics engine as a black box, estimating policy gradients solely from scalar returns. Even with massively parallel GPU simulation~\cite{makoviychuk2021isaac,rudin2021walkinminutes}, whole-body humanoid skills require hours to days of wall-clock training and up to thousands of GPU-hours~\cite{luo2026sonic}. First-order policy optimization directly differentiates the return through the environment's transition dynamics, offering dramatically lower sample complexity in principle. Yet on contact-rich robots, first-order methods have largely failed in practice: backpropagating through stiff contacts produces exploding gradients~\cite{metz2021gradients,suh2022dodiff}, and existing differentiable simulators are orders of magnitude slower than forward-only engines.

Recent humanoid tracking works showed that backpropagating through a learned world model accelerates training relative to PPO~\cite{liao2026supermimic,dao2026instanttrack}. Rather than abandoning physics gradients for learned surrogates, we ask what happens when true environment gradients are made both stable and fast---and how exact adjoints, analytic surrogates, and learned world models compare within a single controlled trainer.

We present \napo\footnote{\napo is pronounced ``Napo'' (or ``Nabla-P-O'').}, a first-order policy optimization method built on three core pillars (\cref{sec:method}): (i) \emph{per-sample clipping}, which clips each sample's action gradient relative to the batch median so that isolated contact spikes cannot dominate the update; (ii) \emph{multi-step action-target regression}, which converts clipped action gradients into fixed target displacements in physical joint units so that each sampled batch supports several regression steps within an automatic \emph{action-space trust region}; and (iii) \emph{ultrafast exact adjoints} via \engine, our GPU simulator that evaluates full MuJoCo contact dynamics and exact step adjoints at the throughput of forward-only simulators. Because \napo's backward pass also accepts learned world models (\napo-WM) and straight-through analytic surrogates (\cref{fig:teaser}), it provides a unified testbed to analyze the gradient spectrum (\cref{sec:axis}).

Our contributions are fourfold:
\begin{itemize}[leftmargin=*,itemsep=0.5pt,topsep=1.5pt]
  \item \textbf{\napo reaches the milestone returns of PPO, SuperMimic, and InstantTrack $9$--$200\times$ sooner while raising the asymptotic ceiling.} Training one policy across all 40 LAFAN1 clips (\cref{fig:hero}, \cref{sec:tracking}), \method reaches the plateau return of a 6-hour PPO run in 1.5\,min ($200\times$ faster than the 5.0\,h at which that run first reaches it), SuperMimic's reported 3-hour score in 5.3\,min ($34\times$ faster), and InstantTrack's plateau in 17\,min vs.\ 2.6\,h ($9\times$ faster). It converges to $0.929\pm0.002$ with zero fallen clips (+4.4\% over InstantTrack, +10.9\% over SuperMimic's official code), reduces blind root drift over $>200$\,s clips by $2.1\times$ and $7.6\times$, and deploys directly from simulation to a physical Unitree G1 without real-world fine-tuning.
  \item \textbf{\engine computes exact MuJoCo contact adjoints $8$--$12\times$ faster and forward steps $5\times$ faster than MuJoCo Warp.} Fused GPU kernels evaluate full MuJoCo contact dynamics and exact step adjoints matching float64 finite differences; final-recipe policies trained in \engine score within 0.001 normalized return when re-evaluated in MuJoCo Warp (\cref{sec:method,sec:tracking}). \engine is open source at \url{https://github.com/sailab-ai/NablaEngine}.
  \item \textbf{Controlled gradient analysis shows that exact contact gradients fail from single-sample heavy tails---not bias---and deliver the strongest learning signal per update.} Under identical policies, critics, and data (\cref{sec:axis}), exploding contact gradients reside in isolated samples ($>10^9\times$ the batch median while the 99.9th percentile stays flat). Once clipped per sample, exact adjoints learn fastest per update; strong world models align with them up to cosine 0.94 and inherit clipping gains, while analytic surrogates help where exact derivatives vanish.
  \item \textbf{A single first-order learner generalizes across 16 control benchmarks and 4 additional humanoid skills within minutes.} With per-task configurations from one small grid, \napo beats PPO in time-to-80\% return on 14 of 16 DeepMind Control and Brax tasks (one fixed configuration does so on 12 of the 13 timed tasks) and exceeds the final return of SHAC and AHAC on every task they were run on (\cref{sec:tasks}). On the Unitree G1, it learns velocity tracking in 3\,min, 2\,kg box lifting in 10\,min, unified joystick-and-motion control in 11\,min, and 10-terrain traversal in 20\,min on a single GPU.
\end{itemize}
\section{Related work}
\label{sec:related}

\paragraph{Zeroth-order policy learning.} GPU simulation~\cite{makoviychuk2021isaac,rudin2021walkinminutes,
mittal2025isaaclab,zakka2025playground,zakka2026mjlab} made rollouts cheap, enabling likelihood-ratio methods such as
PPO~\cite{schulman2017proximal} and off-policy actor-critics such as SAC~\cite{haarnoja2018sac} and
FastTD3~\cite{seo2025fasttd3}. Humanoid motion tracking followed this route from
DeepMimic~\cite{peng2018deepmimic} and PHC~\cite{luo2023perpetual} to BeyondMimic~\cite{liao2026beyondmimic} on the Unitree G1 and SONIC~\cite{luo2026sonic}, which scales PPO tracking to 21k GPU-hours. All treat the simulator as a black box.

\paragraph{First-order policy learning.} Backpropagating through differentiable
simulators~\cite{degrave2019diffphys,hu2020difftaichi,freeman2021brax} suffers from exploding
gradients~\cite{metz2021gradients} and high contact variance~\cite{suh2022dodiff,
suh2022bundled}. PODS~\cite{mora2021pods} regresses policies onto gradient-improved actions; SHAC~\cite{xu2022shac}
introduced short windows with a TD($\lambda$) critic, extended by AHAC~\cite{georgiev2024ahac} and SAPO~\cite{xing2025sapo}. Legged-robot works train through smoothed contact~\cite{schwarke2025deployable}, simplified backward
models~\cite{song2024quadruped}, or smoothed backward contact sensitivity~\cite{kim2026diffcontactlegged}, typically with simplified contact and small batch sizes.

\paragraph{Gradients through learned models.} SVG, MAAC, and Dreamer~\cite{heess2015svg,clavera2020maac,hafner2020dreamer}
optimize policies through learned dynamics, MoSim~\cite{hao2025mosim} predicts long horizons of raw physical state with a learned motion simulator, and PWM~\cite{georgiev2025pwm} argues that learned models smooth the optimization landscape. For humanoid tracking, SuperTrack~\cite{fussell2021supertrack},
SuperMimic~\cite{liao2026supermimic}, and InstantTrack~\cite{dao2026instanttrack} backpropagate through a learned world model. \napo places learned models on one spectrum alongside exact simulator adjoints and analytic surrogates (\cref{sec:axis}).

\begin{table}[t]
\centering
\small
\caption{\textbf{\napo unifies exact, analytic-surrogate, and learned-model gradients to solve both whole-body humanoid tracking and general control benchmarks in minutes.} \checkmark: demonstrated; (\checkmark): one auxiliary skill (velocity tracking); $\times$: not shown; n/a: no tracking.}
\label{tab:related}
\setlength{\tabcolsep}{3pt}
\resizebox{\linewidth}{!}{%
\begin{tabular}{@{}lcccc@{}}
\toprule
Method & Gradient & Motion tracking & Other tasks & Time \\
\midrule
PPO~\cite{schulman2017proximal,liao2026beyondmimic} & likelihood ratio & \checkmark & \checkmark & hours--days \\
SAC~\cite{haarnoja2018sac} & critic (Q) & $\times$ & \checkmark & n/a \\
SHAC, AHAC~\cite{xu2022shac,georgiev2024ahac} & exact (penalty contact) & $\times$ & \checkmark & n/a \\
SuperMimic, InstantTrack~\cite{liao2026supermimic,dao2026instanttrack} & learned model & \checkmark & (\checkmark) & hours \\
\napo (ours) & exact / analytic / learned & \checkmark & \checkmark & minutes \\
\bottomrule
\end{tabular}}
\end{table}

\paragraph{Differentiable simulators.} DiffTaichi and Warp generate adjoint kernels~\cite{hu2020difftaichi,
macklin2022warp}; Brax and MJX~\cite{freeman2021brax,mjx2024} use JAX autodiff; Dojo~\cite{howell2022dojo} differentiates the contact solution via the implicit function theorem, and DiffMJX~\cite{paulus2026diffmjx} differentiates MuJoCo soft contacts~\cite{todorov2012mujoco}. MuJoCo Warp~\cite{mujocowarp2025} provides an adjoint only in an open pull request~\cite{mjwarp_pr1535}. \engine computes exact, deterministic adjoints of MuJoCo's full contact dynamics at the throughput of forward-only GPU simulators.
\section{\napo}
\label{sec:method}

\begin{figure}[t]
\centering
\includegraphics[width=0.80\linewidth]{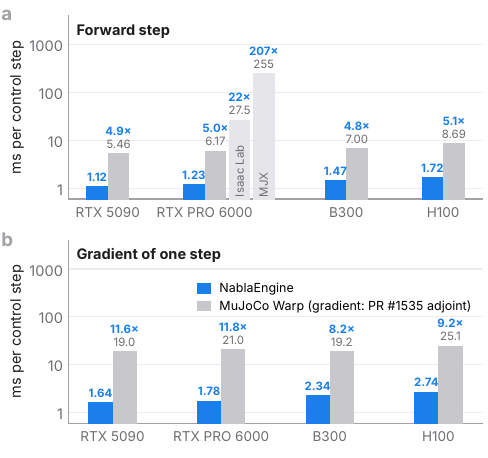}
\caption{\textbf{\engine accelerates Unitree G1 forward simulation by $4.8$--$5.1\times$ and single-step exact gradients by $8.2$--$11.8\times$ over MuJoCo Warp across four GPUs.} Time per 20\,ms control step across 4096 worlds (log scale): (a) forward step, (b) 1-step gradient (forward with tape + backward), with wall-clock time and speedup multiple (blue) above each bar. MuJoCo Warp uses its open adjoint PR~\cite{mjwarp_pr1535}; Isaac Lab and MJX are measured on RTX PRO 6000.}
\label{fig:engine}
\end{figure}

A policy-gradient method estimates $\nabla_\theta \mathbb{E}[R]$ for trajectories generated by policy $\pi_\theta$ under dynamics $s_{t+1} = f(s_t, a_t)$. Whereas zeroth-order methods such as PPO~\cite{schulman2017proximal} weight $\nabla_\theta \log \pi_\theta$ by scalar returns, first-order optimization backpropagates returns through the transition Jacobians. Following SHAC~\cite{xu2022shac}, \napo bounds backpropagation depth by simulating batches of $H$-step windows ($H=64$ on the humanoid, $H=8$ on control benchmarks) from start states sampled from a replay buffer populated by an asynchronous collector. With actions $a_t = \pi_\theta(o_t) + \epsilon_t$ and a TD($\lambda$) critic $V_\phi$ bootstrapping beyond step $H$,
\begin{equation}
  J = \sum_{t<H} \gamma^t r(s_t, a_t) + \gamma^H V_\phi(s_H),
  \label{eq:window-return}
\end{equation}
the backward pass evaluates the adjoint recursion
\begin{equation}
  \frac{\partial J}{\partial a_t} = \frac{\partial r_t}{\partial a_t} + \lambda_{t+1}^{\!\top} \frac{\partial f}{\partial a_t},
  \qquad
  \lambda_t = \frac{\partial r_t}{\partial s_t} + \lambda_{t+1}^{\!\top} \frac{\partial f}{\partial s_t},
  \label{eq:costate}
\end{equation}
to obtain action gradients $g_t = \partial J/\partial a_t$. \napo introduces three components to make environment gradients stable, sample-efficient, and fast.

\paragraph{Per-sample clipping.}
Rigid-body contact is non-differentiable at grazing impacts and geometry transitions, causing rare individual samples in a batch to reach $10^9$--$10^{10}$ times the batch median gradient norm while 99.9\% of the batch remains well-scaled (\cref{sec:axis}). Under global norm clipping, a single outlier dominates the batch norm and rescales all other samples to near zero. \napo instead clips each sample's action gradient relative to the batch median norm:
\begin{equation}
  \tilde g_i = g_i \min\!\left(1, \frac{\kappa\, \operatorname{median}_j \lVert g_j \rVert}{\lVert g_i \rVert}\right).
  \label{eq:clip}
\end{equation}
Because the threshold adapts to the batch median, a single $\kappa=3$ generalizes across tasks whose raw gradient scales differ by orders of magnitude. Our implementation applies \cref{eq:clip} across the batch at each step of the backward pass and once more to the pooled action gradients of the window (\cref{alg:supp-nablapo}); in the final recipe, clipping only the window's action gradients trains equally well (\cref{tab:supp-abl-final}), whereas removing per-sample clipping collapses every seed.

\paragraph{Multi-step action-target regression.}
Differentiating an $H$-step physics window costs far more than evaluating the actor MLP, so a single gradient step per window underutilizes each batch. Yet taking multiple raw gradient steps on a fixed window quickly leaves the neighborhood where its first-order Taylor expansion holds, with step sizes tied to wildly varying gradient norms. To reuse each sampled batch safely, \napo converts the clipped gradients $\tilde g_t$ into fixed target actions $y_t$ and regresses the policy onto them with $M$ minibatch steps ($M=8$ on the humanoid, each on 16{,}384 targets drawn with replacement from the window's $2048\times64=131{,}072$ samples, so the $M$ steps make about one pass over the window in expectation):
\begin{equation}
  y_t = a_t + \delta\, \frac{\tilde g_t}{\operatorname{RMS}(\tilde g)}, \qquad
  \min_\theta \sum_t \lVert \pi_\theta(o_t) - y_t\rVert^2 .
  \label{eq:target}
\end{equation}

\begin{figure}[t]
  \centering
  \includegraphics[width=0.78\linewidth]{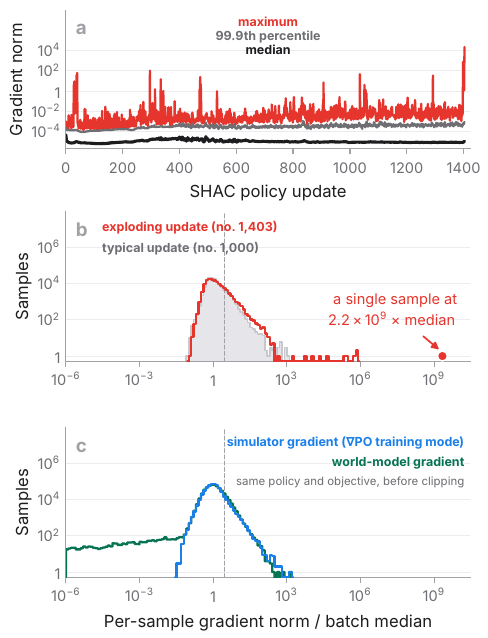}
  \caption{\textbf{Exploding contact gradients reside in single outlier samples, while 99.9\% of the batch remains well-behaved.} Per-sample action-gradient norms of SHAC on exact humanoid tracking gradients (before clipping). (a)~Across updates, the batch median and 99.9th percentile stay flat; only the maximum spikes. (b)~Within one batch of 131{,}072 samples (relative to median): a typical update (no.~1{,}000, grey) vs.\ the exploding update (no.~1{,}403, red) that halts SHAC, where one sample reaches $2.2\times10^9\times$ median. (c)~For an identical policy and 32-step window, pre-clipping per-sample gradient norms of \napo's simulator gradient and a world model's share the same heavy right tail ($\sim 10^3\times$ median); the world model's left tail comes from un-terminated post-fall states (3.4\% of rows). Dashed: clipping threshold ($3\times$ median).}
  \label{fig:grad-tails}
\end{figure}

\begin{figure*}[t]
\centering
\includegraphics[width=0.84\linewidth]{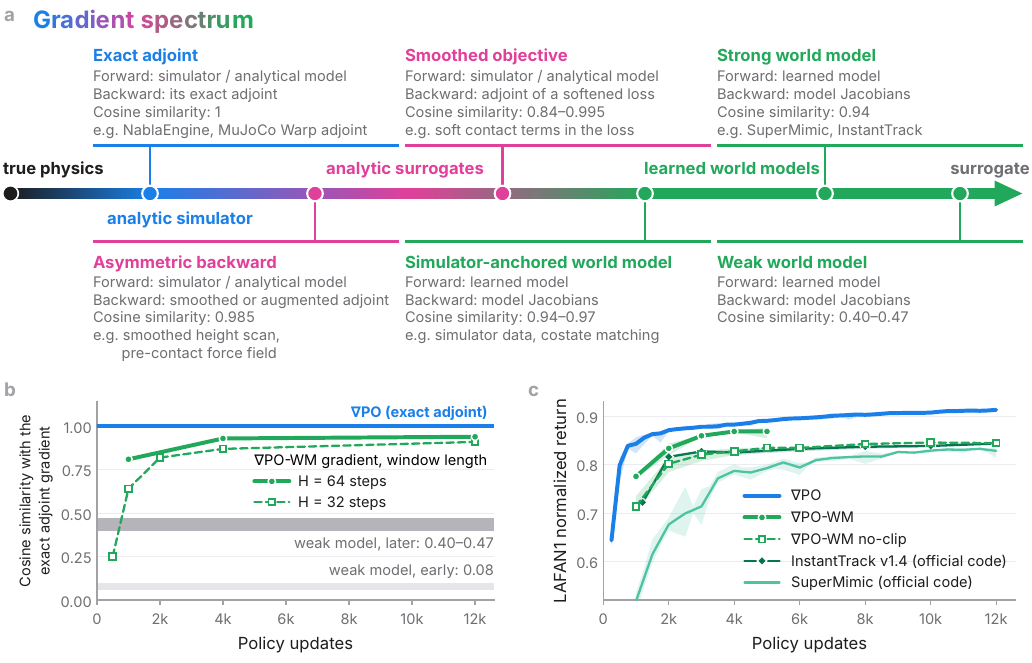}
\caption{\textbf{Exact adjoints deliver the fastest learning per update, while strong world models align with them up to cosine 0.94 and inherit \napo's clipping gains.} (a) The gradient spectrum in \napo: exact simulator adjoints (blue), analytic surrogates with bitwise-exact forward rollouts (pink), and learned world models (green), with median per-sample cosine similarities against the exact action gradient $\partial J/\partial a_t$. (b) Per-sample cosine similarity between world-model and exact gradients over velocity-tracking training: the strong model (\napo-WM) reaches cosine $0.94$ at $H=64$ ($0.91$ at $H=32$); a weak model stays below $0.5$. (c) 40-clip LAFAN1 return vs.\ policy updates under identical evaluation: \napo with exact adjoints (5 seeds) learns fastest per update, ahead of \napo-WM (3 seeds). Removing \napo's window and clipping (\napo-WM no-clip, dashed) drops the curve onto official InstantTrack v1.4 (3 seeds), isolating the gain of \napo's components.}
\label{fig:axis}
\end{figure*}

\paragraph{Action-space trust region.}
Unlike PODS~\cite{mora2021pods}, which regresses onto unnormalized gradient steps, normalizing $\tilde g_t$ by its batch RMS fixes the update radius in physical action units: the batch RMS displacement is $\delta$, and combined with per-sample clipping (\cref{eq:clip}), no sample's per-joint RMS target change exceeds $\sqrt{2}\,\kappa\,\delta$ ($0.13$\,rad), independent of gradient scale (\cref{eq:supp-bound}; without clipping, one outlier moves its target by $\sim 15$\,rad and freezes the rest). As $\pi_\theta(o_t)$ approaches $y_t$ over the $M$ steps, the residual $\pi_\theta(o_t) - y_t$ shrinks automatically, forming an action-space trust region without KL penalties or line search. Because \engine is deterministic, we calibrate $\delta$ directly via 50-update branch probes with $\delta$ scaled by $\{0.5, 1, 2, 4\}\times$ from a checkpoint on fixed start states (\cref{tab:supp-branch}): large radii win early, $1\times$ and $2\times$ tie after 250 updates, and $4\times$ overshoots after 1{,}000 updates; we therefore decay $\delta$ together with the learning rate. Smaller late steps also prevent late-training collapse: a run whose return had begun to drop ($0.886 \to 0.653$) instead climbs to $0.915$ when resumed with the regression optimizer's parameter update scaled by $0.05$ and $\delta$ unchanged.

\paragraph{Ultrafast exact adjoints.}
To remove the simulator bottleneck, \engine evaluates MuJoCo's full contact dynamics via fused per-world GPU kernels and computes the exact adjoint of an entire control step in a single fused kernel matching float64 finite differences. For the Unitree G1 across 4096 worlds on an RTX PRO 6000, a forward step takes $1.23$\,ms and a gradient step takes $1.78$\,ms---$4.8$--$5.1\times$ faster in forward simulation and $8.2$--$11.8\times$ faster in adjoint computation than MuJoCo Warp across four GPU generations (\cref{fig:engine}). For motion tracking, \napo shares the goal encoder-decoder actor~\cite{liao2026supermimic} and EMA policy evaluation~\cite{dao2026instanttrack} of prior trackers, isolating the effect of gradient optimization.
\section{From exact to surrogate gradients}
\label{sec:axis}

\paragraph{Environment gradient interface.}
While \napo uses \engine's exact adjoint by default, the window backward pass in \cref{eq:costate} is modular: holding the policy, critic, replay collector, and evaluator fixed, we can swap the dynamics Jacobians $(\partial f/\partial s_t, \partial f/\partial a_t)$ along a continuous spectrum (\cref{fig:axis}a). At one end is \engine's \emph{exact adjoint}; at the other is a \emph{learned world model} (\napo-WM) trained online on collector transitions to replace both passes of the training window, as in SuperMimic~\cite{liao2026supermimic} and InstantTrack~\cite{dao2026instanttrack}. Between them lie \emph{analytic surrogates}, which keep the forward rollout bitwise exact while modifying the backward VJP or smoothing the training objective.

\paragraph{Exact contact gradients are heavy-tailed, not biased.}
\engine's training adjoint is bitwise identical to the exact discrete derivative except at two degenerate geometric events where the analytical derivative is unbounded (self-contact core distance $<1$\,mm and capsule pairs within $2^\circ$ of parallel); along a trajectory, per-sample gradient norms of the two modes have a Spearman rank correlation of $0.9999$. Exact contact gradients fail in standard trainers not from directional bias, but because stiff contacts create extreme single-sample outliers ($10^9$--$10^{10}\times$ the batch median). Under global norm clipping as in SHAC~\cite{xu2022shac}, all three humanoid tracking seeds trip their exploding-gradient guard within 5\,min at a return of $0.38$, while the batch median, 99th, and 99.9th percentiles stay flat (\cref{fig:grad-tails}a,b). Disabling the guard so exploding batches are globally clipped still leaves each update dominated by the single extreme sample, causing SHAC to plateau at $0.817$ after 3\,h---well below the $0.866$ that \method passes in 5.3\,min (\cref{tab:levels}). Once per-sample clipping (\cref{eq:clip}) removes these isolated spikes, the exact adjoint delivers the highest learning progress per update of any gradient source tested (\cref{fig:axis}c).

\begin{figure*}[t]
  \centering
  \includegraphics[width=0.84\linewidth]{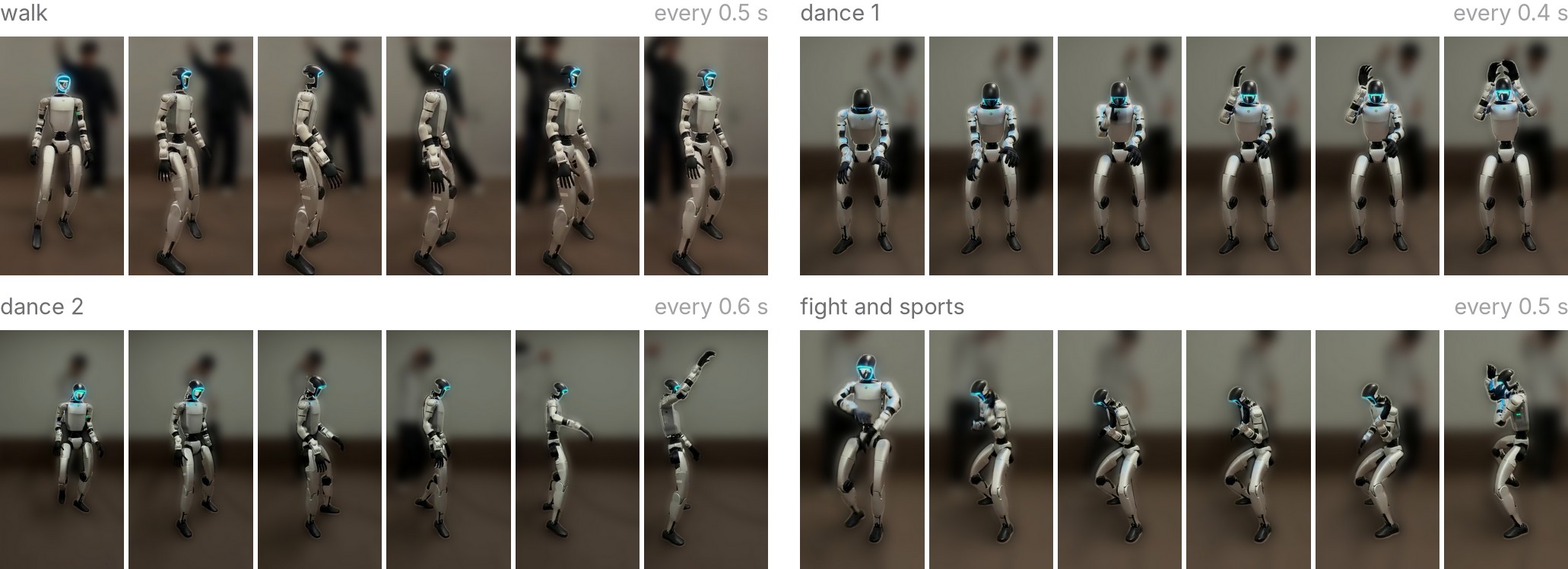}
  \caption{\textbf{A \method policy deploys directly from simulation to a physical Unitree G1 without real-world fine-tuning, executing agile whole-body motions without a fall.} Checkpoints from earlier recipes than the final one (\cref{supp:mpjpe-real}). Six consecutive frames ($0.4$--$0.6$\,s apart) from four LAFAN1 clips (walk, two dances, and fight/sports sequences) executed at 50\,Hz with a slack safety strap.}
  \label{fig:real}
\end{figure*}

\paragraph{World-model gradients track model fidelity and inherit \napo's gains.}
A learned world model avoids differentiating through contact solvers and has globally Lipschitz-bounded Jacobians. Yet bounded Jacobians still produce heavy-tailed multi-step gradients: on the same policy and 32-step window, the world model's per-sample gradient norms share the same $\sim 10^3\times$ right tail as \engine's exact gradient (\cref{fig:grad-tails}c), explaining why per-sample clipping benefits world-model training as well. Across training, a strong world model (InstantTrack's structured architecture) aligns closely with the exact gradient, reaching a median per-sample cosine of $0.94$ (\cref{fig:axis}b), whereas a weaker model stays below $0.5$ with $13$--$30\%$ opposite-sign samples. On 40-clip LAFAN1 tracking at 4k updates (\cref{fig:axis}c), exact gradients learn fastest ($0.883$), followed by the strong world model with \napo's 64-step window and per-sample clipping (\napo-WM, $0.869$), and finally InstantTrack's unclipped 32-step settings (\napo-WM no-clip, $0.828$, matching official InstantTrack per update). On the same A40 GPU, \napo-WM reaches $0.86$ in 25\,min vs.\ 68\,min for InstantTrack. \napo retains \napo-WM for tasks without a differentiable simulator, while achieving peak efficiency with exact adjoints.

\paragraph{Analytic surrogates supply signal where exact derivatives vanish.}
Where a reward or sensor has zero derivative before contact or across flat geometry, \engine can inject an analytic surrogate into the backward pass while keeping forward states, rewards, and evaluations bitwise exact. For a robot--object pair at distance $d$ with normal $n$, \engine adds a straight-through pre-contact impulse
\begin{equation}
  \Delta\dot q = M^{-1} J_c^{\!\top} n \; e^{-d/w}\,(k - c\,\dot d)\,\Delta t \quad \text{with} \quad
  \mathrm{value}(\Delta\dot q) = 0,
  \label{eq:surrogate}
\end{equation}
whose forward value is zero but whose Jacobian transmits a decaying approach gradient before impact (with width $w$ annealed). Similarly, replacing the zero derivative of a piecewise-flat terrain height scan with the VJP of a Gaussian-smoothed scan matches central finite differences of the smoothed scan at cosine $0.985$ and exposes step edges. On humanoid box pushing across randomized box sizes, the backward-only impulse (\cref{eq:surrogate}) lowers final box position error from $4.6$ to $3.8$\,cm ($p\approx 0.03$), while smoothing the contact objective cuts box error from $6.5$ to $2.9$\,cm across 6 seeds.
\section{Motion tracking: \method}
\label{sec:tracking}

\begin{table}[t]
\centering
\small
\caption{\textbf{\method reaches the plateau of a 6\,h PPO run $200\times$ sooner, SuperMimic's reported 3\,h score $34\times$ sooner, and InstantTrack's plateau $9\times$ sooner, and achieves the highest asymptotic plateau (0.929) on 40-clip LAFAN1 tracking.} Asymptotic plateau return and training minutes on one RTX PRO 6000 GPU to reach four thresholds ($>t$: not reached within $t$ minutes). At the lowest threshold InstantTrack is as fast as \method; at $0.866$ \method is $6.3\times$ faster than InstantTrack.}
\label{tab:levels}
\setlength{\tabcolsep}{3.5pt}
\resizebox{\linewidth}{!}{%
\begin{tabular}{@{}lccccc@{}}
\toprule
Method & Plateau & 0.721 & 0.866 & 0.890 & 0.929 \\
\midrule
\method (5 seeds) & \textbf{0.929}\,$\pm$\,0.002 & \textbf{1.5} & \textbf{5.3} & \textbf{17.2} & \textbf{78} \\
SuperMimic, official code (3 seeds) & 0.838\,$\pm$\,0.021 & 16.1 & $>$190 & $>$190 & $>$190 \\
InstantTrack v1.4 (3 seeds) & 0.890\,$\pm$\,0.001 & \textbf{1.5} & 33.2 & 155 & $>$180 \\
PPO, BeyondMimic (1 seed, 6\,h) & 0.721 & 298 & $>$360 & $>$360 & $>$360 \\
PPO, BeyondMimic (3 seeds, 48\,h)\textsuperscript{a} & 0.782\,$\pm$\,0.005 & 266 & $>$816 & $>$816 & $>$816 \\
SHAC, guard relaxed (3 seeds) & 0.817\,$\pm$\,0.002 & 27 & $>$180 & $>$180 & $>$180 \\
SHAC, published guard (3 seeds) & \multicolumn{5}{l}{guard stops training within 5\,min (0.38)} \\
\bottomrule
\end{tabular}}
\par\smallskip\raggedright\footnotesize
\textsuperscript{a}\,48\,h of training on RTX 3090 (one seed finished on an A40); times converted to the RTX PRO 6000 by update count, so the runs span 13.6--16.8\,h.
\end{table}

\begin{table*}[t]
\centering
\caption{\textbf{\napo reaches 80\% of peak return faster than PPO on 14 of 16 control benchmarks, achieves the highest final return on 13 of 16, and exceeds the final return of SHAC on all 16 and of AHAC on all 15 tasks it was run on.} Final return in the canonical simulator (mean $\pm$ s.d.\ across seeds) and V100 training minutes to 80\% of $R^*$ (best mean return among single-configuration methods; median over seeds; \notreached: $<2$ seeds reach $0.8\,R^*$; \notrun: not run). \napo: per-task configuration (3--5 seeds; \cref{tab:supp-dmc-configs}); \napo$^\dagger$: one identical configuration across all tasks (5 seeds). PPO and SAC are run in \engine (NE) and MuJoCo Playground (orig.); SHAC and AHAC use published settings (3 seeds).}
\label{tab:dmc}
\setlength{\tabcolsep}{2.6pt}
\renewcommand{\arraystretch}{0.84}
\scriptsize
\begin{tabular}{@{}lcccccccc@{\hspace{10pt}}ccc@{}}
\toprule
& \multicolumn{8}{c}{Final return} & \multicolumn{3}{c}{Minutes to 80\%} \\
\cmidrule(lr){2-9}\cmidrule(l){10-12}
Task & \napo & \napo$^\dagger$ & PPO NE & PPO orig. & SAC NE & FastTD3 & SHAC & AHAC & \napo & PPO & SAC \\
\midrule
acrobot-swingup & \textbf{250$\pm$21} & 237$\pm$48 & 227$\pm$59 & 151$\pm$13 & 52$\pm$52 & 62$\pm$33 & 86$\pm$46 & 51$\pm$43 & 7.4 & 10.6 & \notreached \\
acrobot-swingup-sparse & \textbf{104$\pm$34} & 77$\pm$12 & 0 & 11$\pm$8 & 0 & 5$\pm$6 & 2$\pm$1 & 2$\pm$2 & 13.6 & \notreached & \notreached \\
cartpole-swingup & \textbf{883} & 883 & 800$\pm$75 & 785$\pm$103 & 866 & 314$\pm$113 & 419$\pm$200 & 224$\pm$49 & 1.3 & 4.6 & 2.1 \\
cartpole-swingup-sparse & \textbf{675$\pm$139} & 0 & 0 & 0 & 539$\pm$467 & 69$\pm$60 & 0 & 2$\pm$4 & 9.7 & \notreached & 7.1 \\
cheetah-run & \textbf{919$\pm$4} & 917$\pm$5 & 79$\pm$60 & 177$\pm$250 & 907$\pm$4 & 908$\pm$4 & 373$\pm$83 & 417$\pm$52 & 1.5 & 10.9 & 10.5 \\
finger-spin & 955$\pm$11 & 945$\pm$21 & 762$\pm$230 & 467$\pm$19 & \textbf{962$\pm$40} & 506$\pm$143 & 0 & 3$\pm$5 & 6.8 & 14.5 & 10.7 \\
hopper-hop & \textbf{395$\pm$64} & 333$\pm$21 & 11$\pm$15 & 2 & 210$\pm$39 & 2$\pm$1 & 2$\pm$2 & 8$\pm$14 & 5.3 & \notreached & \notreached \\
hopper-stand & \textbf{939$\pm$16} & 839$\pm$252 & 101 & 54$\pm$67 & 761$\pm$326 & 120$\pm$55 & 4$\pm$2 & 4$\pm$1 & 1.8 & \notreached & 17.5 \\
pendulum-swingup & \textbf{792$\pm$4} & 781$\pm$15 & 785$\pm$12 & 790$\pm$2 & 448$\pm$500 & 410$\pm$393 & 33$\pm$28 & 31$\pm$27 & 1.1 & 1.6 & \notreached \\
reacher-hard & 974$\pm$4 & 968$\pm$3 & 978$\pm$3 & 924$\pm$74 & \textbf{979} & 974$\pm$10 & 221$\pm$283 & 184$\pm$179 & 1.3 & 1.9 & 1.3 \\
walker-run & \textbf{904$\pm$50} & 865$\pm$45 & 686$\pm$139 & 384$\pm$335 & 816$\pm$3 & 453$\pm$47 & 34$\pm$9 & 27$\pm$7 & 4.1 & 23.1 & 15.8 \\
walker-walk & \textbf{981$\pm$1} & 980$\pm$2 & 976$\pm$1 & 970 & 976$\pm$2 & 612$\pm$97 & 43$\pm$12 & 37$\pm$16 & 4.8 & 6.6 & 3.9 \\
\midrule
Brax ant & \textbf{7019$\pm$666} & 650$\pm$176 & 4174$\pm$244 & 4113$\pm$363 & 4220$\pm$471 & \notrun & -262$\pm$402 & \notrun & 18.6 & \notreached & \notreached \\
Brax halfcheetah & \textbf{17689$\pm$235} & 15623$\pm$994 & 5123$\pm$1296 & 4983$\pm$1489 & 11780$\pm$644 & \notrun & -197$\pm$442 & 5$\pm$331 & 6.8 & \notreached & \notreached \\
Brax hopper & 2716$\pm$518 & 2716$\pm$518 & \textbf{3589$\pm$66} & 3528$\pm$131 & 3358$\pm$229 & \notrun & 42$\pm$6 & 1063$\pm$63 & \notreached & 4.0 & 28.8 \\
Brax walker2d & \textbf{6725$\pm$203} & 5918$\pm$656 & 6641$\pm$591 & 5108$\pm$1316 & 5563$\pm$405 & \notrun & 213$\pm$188 & 243$\pm$59 & 14.7 & 7.2 & 40.0 \\
\bottomrule
\end{tabular}
\end{table*}

\begin{figure*}[t]
\centering
\includegraphics[width=0.80\linewidth]{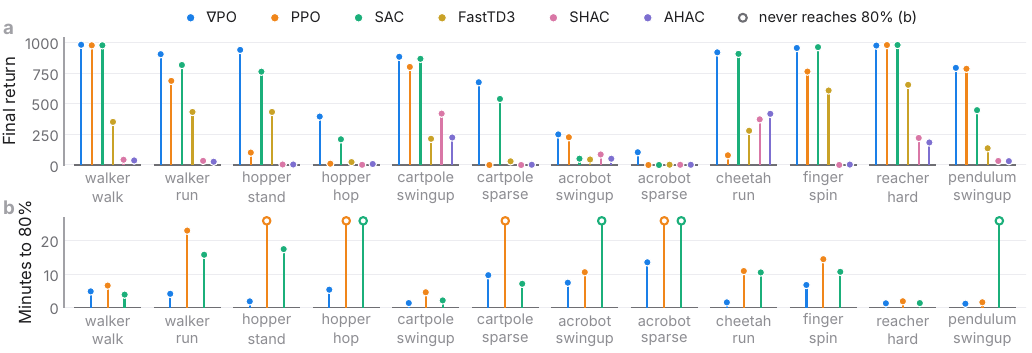}
\caption{\textbf{Across all 12 DeepMind Control tasks, \napo matches or exceeds the asymptotic return of zeroth- and first-order baselines in a fraction of the training time.} (a) Final checkpoint return (seed mean). (b) Training minutes until the median seed reaches $0.8\,R^*$ (\cref{tab:dmc}); hollow dots mark runs that never reach it.}
\label{fig:dmc}
\end{figure*}

Whole-body humanoid motion tracking is the most demanding testbed for first-order policy optimization and where the strongest surrogate-gradient baselines exist. We denote \napo applied to whole-body tracking as \method.

\paragraph{Setup and baselines.}
One policy tracks all 40 LAFAN1 clips~\cite{harvey2020robust} retargeted to the 29-DoF Unitree G1~\cite{lafan1retarget2025} (2.45\,h of motion), rolled out deterministically from frame 0 without noise. Following SuperMimic~\cite{liao2026supermimic}, we report full-clip normalized return (tracking reward minus action-rate and joint-limit penalties, summed over the clip and normalized by its maximum, so an early fall forfeits the remaining frames, averaged across all 40 clips). A run's plateau is its mean return over the final 30\,min across seeds; time-to-level is the first crossing of the seed-mean curve from the first training update (excluding startup compilation and evaluation). We compare against: (i) \textbf{PPO}~\cite{schulman2017proximal} (BeyondMimic configuration~\cite{liao2026beyondmimic}) in \engine; (ii) \textbf{SHAC}~\cite{xu2022shac} in \engine, with its published exploding-gradient guard and with the guard relaxed for global norm clipping; (iii) \textbf{SuperMimic}~\cite{liao2026supermimic} via its official codebase (using the paper's actor observations, history length, and 40-clip dataset; \cref{supp:tracking-setup}); and (iv) \textbf{InstantTrack v1.4}~\cite{dao2026instanttrack}, a fast open-source world-model tracker with EMA policy averaging.

\paragraph{Controlled comparison.}
All methods executed in our benchmark share the \textbf{identical Unitree G1 MJCF model, 40 LAFAN1 reference motions, observation and action spaces, reward and termination functions, single RTX PRO 6000 GPU budget, and deterministic MuJoCo Warp evaluation protocol} (re-scoring the same \method checkpoints in \engine changes the 5-seed plateau mean by at most $0.001$; \cref{tab:supp-sim2sim}).

\begin{figure*}[t]
  \centering
  \includegraphics[width=0.73\linewidth]{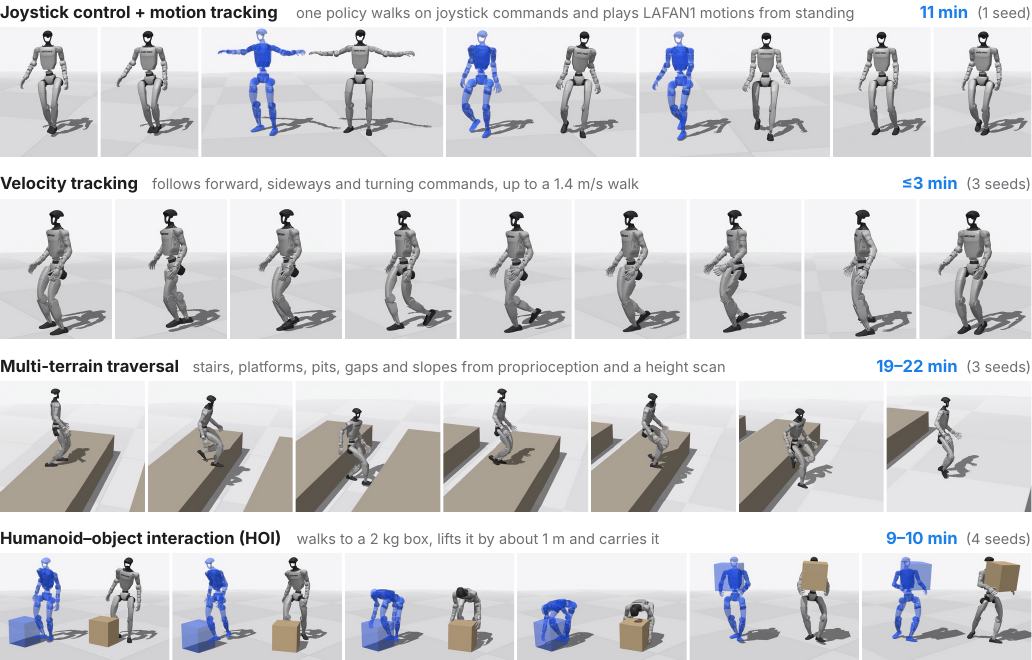}
  \caption{\textbf{One first-order learner trains diverse whole-body humanoid skills in 3 to 22 minutes on a single GPU.} Rollout frame sequences for unified joystick-and-motion control (11\,min), omnidirectional velocity tracking ($\le 3$\,min), 10-type terrain traversal (19--22\,min), and 2\,kg box lifting and carrying (10\,min; blue ghosts denote reference motions).}
  \label{fig:tasks}
\end{figure*}

\begin{table}[t]
\centering
\small
\caption{\textbf{\napo acquires four distinct whole-body humanoid skills on a single GPU in 3 to 22 minutes.} Minutes from the first training update to reach each task's success criterion (seed count in parentheses) alongside representative prior training budgets.}
\label{tab:tasks}
\setlength{\tabcolsep}{3pt}
\resizebox{\linewidth}{!}{%
\begin{tabular}{@{}llcl@{}}
\toprule
Task & Criterion & \napo (min) & Prior reference \\
\midrule
Joystick + motion tracking & 0 falls, med.\ err.\ $\le$ 5\,cm/s; $\le$5\% clip falls\textsuperscript{a} & \textbf{11} (1) & none without distillation \\
Velocity tracking & 0 falls; med.\ err.\ $\le$ 0.05\,m/s, 0.1\,rad/s & \textbf{$\le$3} (3) & G1 flat PPO: $<$30\,min (2$\times$4090)~\cite{zakka2025playground} \\
Multi-terrain traversal & $\ge$99\% of lanes crossed (10 types $\times$ 10 levels) & \textbf{19--22} (3) & Quadruped PPO: $\sim$20\,min~\cite{rudin2021walkinminutes} \\
HOI: 2\,kg box lift & strict success $\ge$ 0.9 (no fall, box err.\ $<$0.2\,m) & \textbf{9--10} (4) & Multi-clip HOI PPO: $\sim$24\,h $\times$ 8 GPUs~\cite{lessmimic2026} \\
\bottomrule
\end{tabular}}
\par\smallskip\raggedright\footnotesize
\textsuperscript{a}\,Evaluated on 33 standing velocity commands and upright LAFAN1 windows; full sweep ($-0.8$ to $1.2$\,m/s forward, $\pm0.5$\,m/s lateral, $\pm1.2$\,rad/s yaw) reaches zero falls at 113\,min. GPUs: single A100 (joystick), RTX PRO 6000 (velocity), A40 (terrain), H200 (HOI).
\end{table}

\paragraph{Minutes instead of hours.}
\Cref{tab:levels} and \cref{fig:hero} compare wall-clock training times across four return milestones: $0.721$ (the plateau of a 6\,h PPO run), $0.866$ (SuperMimic's reported 3\,h score~\cite[\S4.1]{liao2026supermimic}), $0.890$ (InstantTrack's plateau), and $0.929$ (\method's plateau). \method reaches the 6-hour PPO plateau in \textbf{1.5\,min} ($200\times$ faster than the 5.0\,h at which that run first reaches it; its best checkpoint scores $0.744$ and its final one $0.697$; three further PPO seeds trained for 48\,h first reach $0.721$ after 4.4\,h at the RTX PRO 6000 update rate and end at $0.782\pm0.005$, never reaching $0.82$), SuperMimic's 3-hour score in \textbf{5.3\,min} ($34\times$ faster), and InstantTrack's plateau in \textbf{17.2\,min} vs.\ 2.6\,h ($9\times$ faster). InstantTrack is equally fast to the lowest milestone ($1.5$\,min) but needs $33$\,min for $0.866$, $6.3\times$ longer than \method. SuperMimic's official code reaches $0.82$ in 45.5\,min (vs.\ 2.5\,min for \method) and plateaus at $0.838\pm0.021$. \method converges to \textbf{$0.929\pm0.002$} at 78\,min, with all 5 seeds first at zero falls on all 40 clips at 5.3\,min and at zero falls at every checkpoint from 57\,min onward. Over the 30 clips longer than 200\,s, evaluated blind (without observing root position relative to the reference), \method's mean global root drift is just \textbf{$0.15\pm0.02$\,m} (5 seeds, 0 falls), vs.\ $0.32\pm0.02$\,m for InstantTrack ($2.1\times$ higher) and $1.17\pm0.56$\,m for SuperMimic's official code ($7.6\times$ higher). Starting every training window from the live rollout state, without the replay of collected start states, speeds training further: in a later run of this variant of the recipe (3 seeds; minutes from the update count at its measured 0.238\,s per update on the RTX PRO 6000), the seed-mean curve passes the 6-hour PPO level after 1.2\,min, SuperMimic's reported score after 4.4\,min and InstantTrack's plateau after 8.3\,min, and two seeds trained to 24k updates on that GPU level off at $0.938\pm0.004$.

\paragraph{Engine speed vs.\ gradient efficiency.}
Training \method with the MuJoCo Warp adjoint instead of \engine produces nearly identical learning curves per update (within $0.02$), while \engine completes the run $4.7\times$ faster in wall-clock time (\cref{fig:supp-operator}). Conversely, running PPO on \engine accelerates each PPO iteration by $2.9\times$ over stock MuJoCo Warp, yet PPO still needs 5.0\,h to reach the return \method achieves in 1.5\,min---confirming that the $200\times$ speedup comes from first-order environment gradients rather than simulator speed alone.

\paragraph{Direct deployment on the physical robot.}
A \method tracker deploys directly from simulation to a physical Unitree G1 at 50\,Hz, with no real-world data, system identification, or fine-tuning. Two checkpoints from recipes earlier than the final one, trained for 9.8\,min and for 3.0\,h on one GPU, each ran the same seven trials over five dynamic LAFAN1 clips (walking, three dances, and a fighting sequence) without a fall (\cref{fig:real}). The telemetry of the 3.0\,h policy shows $3.9^\circ$ frame-weighted joint RMS error (\cref{tab:supp-real}).
\section{One method, many tasks}
\label{sec:tasks}

\paragraph{Standard control benchmarks.}
To test generality without task-specific engineering, we evaluate \napo with one fixed configuration (\napo$^\dagger$: Adam, $H=8$ window, 512 environments, Gaussian noise $\sigma=0.2$, actor EMA, $10^8$ steps) alongside per-task tuned \napo across 12 DeepMind Control tasks from MuJoCo Playground~\cite{zakka2025playground} and 4 Brax locomotion tasks~\cite{freeman2021brax}. All methods train in \engine and are evaluated over 10 deterministic episodes in the canonical simulator (\texttt{dm\_control} 1.0.51 or Brax MJX). We compare against PPO and SAC in both \engine and Playground's native stack, FastTD3~\cite{seo2025fasttd3}, and SHAC~\cite{xu2022shac} and AHAC~\cite{georgiev2024ahac} under their published settings (\cref{tab:dmc}, \cref{fig:dmc}).

\napo reaches $0.8\,R^*$ faster than PPO on \textbf{14 of the 16 tasks} and achieves the highest final return on \textbf{13 of 16}. Even with a single un-tuned hyperparameter set across all tasks, \napo$^\dagger$ has a higher final return than PPO NE on 10 of 16 tasks; in dedicated single-V100 timing runs (including compilation), \napo$^\dagger$ reaches $0.8\,R^*$ ahead of PPO on 12 of the 13 DMC and HalfCheetah tasks (\cref{tab:supp-dmc-timed}). Earlier first-order methods fall far behind: SHAC on all 16 tasks and AHAC on all 15 it was run on. Extending training to $4\times 10^8$ steps (3 seeds) further boosts \napo to $484\pm36$ on \texttt{acrobot-swingup}, $258\pm136$ on \texttt{acrobot-swingup-sparse} ($H=128$), and $497\pm83$ on \texttt{hopper-hop}.

\paragraph{Whole-body humanoid skills.}
Using the same clipping and action-target regression, \napo trains the Unitree G1 on four further skills in 3--22 minutes on one GPU (\cref{tab:tasks}, \cref{fig:tasks}; specifications in \cref{supp:skills}):
(i) \emph{Unified joystick and motion tracking}: one policy trained from scratch without distillation that walks on omnidirectional velocity commands and switches into LAFAN1 clips via a mode bit (11\,min to criterion; full sweep without falls at 113\,min);
(ii) \emph{Velocity tracking}: following planar linear and yaw velocity commands up to a $1.4$\,m/s walk with domain randomization ($\le 3$\,min, 3 seeds);
(iii) \emph{Multi-terrain traversal}: crossing 10 terrain types across 10 levels (23\,cm stairs, 70\,cm gaps, 45\,cm platforms/pits, stepping blocks, $70\%$ slopes) from proprioception and a $16\times 9$ height scan (19--22\,min to $\ge 99\%$ completion); and
(iv) \emph{Humanoid--object interaction}: walking to a 2\,kg box, lifting it by $\sim 1$\,m, and carrying it along a retargeted trajectory with object state observed (9--10\,min to $\ge 90\%$ strict success across 4 seeds, converging to $99\pm 1\%$ with $2.6$\,cm mean box error).
\section{Conclusion}
\label{sec:conclusion}

\napo demonstrates that first-order policy optimization---stabilized by per-sample gradient clipping and multi-step action-target regression, and accelerated by \engine's fused exact contact adjoints---collapses whole-body robot policy learning from hours or days into minutes on a single GPU. Comparing exact adjoints, straight-through analytic surrogates, and learned world models in one trainer shows that exact simulator gradients deliver the strongest learning signal per update once single-sample contact outliers are clipped, while analytic surrogates supply signal where exact derivatives vanish and world models serve tasks without a differentiable simulator.

\paragraph{Limitations.}
\engine currently targets articulated robot models such as the Unitree G1 with manipulated objects and standard control benchmarks rather than arbitrary scenes. Because pathwise gradients carry zero signal before the first non-zero reward, sparse-reward tasks rely on critic bootstrapping and longer windows (\eg $H=128$ on \texttt{cartpole-swingup-sparse}, solved in 4 of 5 seeds). Finally, while 40-clip motion tracking deploys directly from simulation to a physical Unitree G1 without real-world fine-tuning, our terrain and object-interaction policies use height-scan or object-state observations in simulation; extending them to onboard hardware perception is left to future work.

{
    \small
    \bibliographystyle{ieeenat_fullname}
    \bibliography{refs}
}

\clearpage
\newgeometry{letterpaper,textwidth=6.5in,textheight=9in,top=1in,headsep=0.2in,footskip=0.4in,onecolumn}
\onecolumn
\appendix

\noindent The appendices give the full algorithm, component ablations, engine details, and task specifications.
\begin{itemize}[leftmargin=*,itemsep=1pt,topsep=2pt]
  \item \textbf{\Cref{supp:method}}: pseudocode (\cref{alg:supp-nablapo}), notation (\cref{tab:supp-notation}), window objective and backward recursion (\cref{eq:supp-costate}), the RMS definition and the $\sqrt{2}\,\kappa\,\delta$ target bound (\cref{eq:supp-bound}), and branch-probe calibration of $\delta$ (\cref{tab:supp-branch}).
  \item \textbf{\Cref{supp:ablations}}: component ablations of the final recipe (\cref{tab:supp-abl-final}) and of two earlier recipes (\cref{tab:supp-abl-res,tab:supp-abl-abs,tab:supp-sensitivity}, \cref{fig:supp-ablations,fig:supp-tails-depth}), world-model horizon and clipping ablation with gradient cosines (\cref{tab:supp-wm-ablation,tab:supp-cosine}), and per-update cost on two engines (\cref{fig:supp-operator}, \cref{tab:supp-cost}).
  \item \textbf{\Cref{supp:engine}}: \engine's model, kernel layout and IFT adjoint, timing (\cref{tab:supp-engine-fwd}, \cref{fig:supp-engine-split}), and accuracy against float64 finite differences (\cref{tab:supp-acc-overall,tab:supp-acc-grad-groups}).
  \item \textbf{\Cref{supp:tracking}}: reward, observations, domain randomization and hyperparameters (\cref{tab:supp-reward,tab:supp-hyper}), full LAFAN1 results (\cref{tab:supp-plateaus,tab:supp-lafan-full}), cross-simulator re-scoring and SuperMimic runs (\cref{tab:supp-sim2sim,tab:supp-supermimic}), MPJPE, success and drift (\cref{tab:supp-robustness,tab:supp-drift}), and physical-robot deployment (\cref{tab:supp-real}).
  \item \textbf{\Cref{supp:dmc}}: per-task configurations, selection grid, and timing (\cref{tab:supp-dmc-configs,tab:supp-dmc-grid,tab:supp-dmc-timed}).
  \item \textbf{\Cref{supp:skills}}: specifications of the additional humanoid skills and surrogate-gradient comparisons (\cref{tab:supp-hoi}).
  \item \textbf{Video and code}: videos of simulated rollouts and physical-robot trials (\cref{supp:mpjpe-real}) are on the project page (\projecturl); the code of \engine is at \url{https://github.com/sailab-ai/NablaEngine}.
\end{itemize}

\section{Algorithm Details}
\label{supp:method}

\subsection{Computational Graph and Pseudocode}
\label{supp:comp-graph}

\begin{algorithm}[t]
  \footnotesize
  \caption{One \napo (\method) policy update.}
  \label{alg:supp-nablapo}
  \DontPrintSemicolon
  \KwIn{Actor $\pi_\theta$, EMA actor $\pi_{\bar\theta}$, critic $V_\phi$, FIFO replay buffer $\mathcal{D}$, asynchronous collector, window length $H$, clip factor $\kappa=3$, target RMS radius $\delta$, regression steps $M$ (hyperparameters in \cref{tab:supp-hyper}).}
  \tcp{1. Asynchronous collector (forward-only)}
  Step collector environments with $a_t = \pi_\theta(o_t) + \epsilon_t^{\text{col}}$ ($\sigma_{\text{col}}=0.1$); sample episode resets from return-weighted bins; push visited states to $\mathcal{D}$\;
  \tcp{2. Differentiable $H$-step window rollout on \engine}
  Sample a batch of $B$ window start states $s_0 \sim \mathcal{D}$\;
  \For{$t = 0, \dots, H-1$}{
    Evaluate observation $o_t = \mathrm{obs}(s_t)$ and noisy action $a_t = \pi_\theta(o_t) + \epsilon_t$, $\epsilon_t \sim \mathcal{N}(0, \sigma^2 I)$\;
    Step simulator $s_{t+1}^{\text{raw}} = f(s_t, a_t)$ (forward pass writing $23.6$\,kB tape per world)\;
    Compute per-step training reward $r_t = r_{\text{task}}(s_t, a_t) + r_{\text{pen}}(s_t)$ (\cref{eq:supp-term-pen})\;
    \eIf{fall termination $d_t = 1$ at $s_{t+1}^{\text{raw}}$}{
      Bootstrap $0$ at $s_{t+1}^{\text{raw}}$; reset environment to reference state $s_{t+1} = \mathrm{stop\_grad}(s_{\text{reset}})$ and continue window\;
    }{
      $s_{t+1} = s_{t+1}^{\text{raw}}$\;
    }
  }
  Evaluate terminal bootstrap $V_\phi(s_H)$ on surviving environments at $t=H$\;
  \tcp{3. Closed-loop backward pass with in-backward per-sample clipping}
  Initialize terminal state cotangent $\lambda_H = \gamma^H (1 - d_{H-1}) \nabla_{s_H} V_\phi(s_H)$\;
  \For{$t = H-1, \dots, 0$}{
    If $d_t = 1$ (reset occurred after step $t$), set $\lambda_{t+1} \leftarrow 0$ (no gradient flows across episode resets)\;
    Compute raw action cotangent via step VJP: $g_t = \gamma^t \frac{\partial r_t}{\partial a_t} + \lambda_{t+1}^{\!\top} \frac{\partial f(s_t, a_t)}{\partial a_t}$\;
    \textbf{In-backward per-sample clip} across the $B$ environments at step $t$ ($\varepsilon = 10^{-8}$):
    $\tilde g_{t,i} = g_{t,i} \min\!\bigl(1,\, \frac{\kappa\,\operatorname{median}_j \lVert g_{t,j}\rVert_2 + \varepsilon}{\lVert g_{t,i}\rVert_2 + \varepsilon}\bigr)$\;
    Propagate state cotangent through both direct dynamics and closed-loop policy feedback:
    $\lambda_t = \gamma^t \frac{\partial r_t}{\partial s_t} + \Bigl(\frac{\partial \pi_\theta(o_t)}{\partial s_t}\Bigr)^{\!\top}\!\tilde g_t + \lambda_{t+1}^{\!\top} \frac{\partial f(s_t, a_t)}{\partial s_t}$\;
  }
  \tcp{4. Window-level clipping and RMS-normalized action targets}
  Pool all $N = B \times H$ live window-step cotangents $\{g_i\}_{i=1}^N$ and apply \cref{eq:supp-clip} over the window to obtain $\tilde g_i$\;
  Compute batch-and-joint RMS norm $\operatorname{RMS}(\tilde g) = \bigl(\frac{1}{Nd}\sum_{i=1}^N \lVert \tilde g_i \rVert_2^2 + \varepsilon\bigr)^{1/2}$\;
  Construct fixed action targets $y_t = a_t + \delta\,\tilde g_t / \operatorname{RMS}(\tilde g)$ (\cref{eq:supp-target})\;
  \tcp{5. Multi-step target regression and critic update}
  \For{$m = 1, \dots, M$}{
    Sample minibatch of $(o_t, y_t)$ pairs; update $\theta$ on $\mathcal{L}_{\text{reg}}(\theta) = \frac{1}{|\mathcal{B}|}\sum_{(o_t,y_t)\in\mathcal{B}} \lVert \pi_\theta(o_t) - y_t \rVert_2^2$\;
  }
  Update EMA actor $\bar\theta \leftarrow \beta_n \bar\theta + (1 - \beta_n)\theta$ with $\beta_n = \min(0.999, \frac{1+n}{10+n})$\;
  Update critic $V_\phi$ by regressing onto $\mathrm{TD}(\lambda_{\text{TD}})$ targets computed on pre-transition window states\;
\end{algorithm}

\begin{table}[t]
\centering
\scriptsize
\caption{\textbf{Notation.}}
\label{tab:supp-notation}
\setlength{\tabcolsep}{3.5pt}
\resizebox{\linewidth}{!}{%
\begin{tabular}{@{}lll@{}}
\toprule
Symbol & Domain & Meaning \\
\midrule
$\lambda_t$ & $\mathbb{R}^{\dim(\mathcal{S})}$ & State adjoint (costate) $\partial J / \partial s_t$ at step $t$ (\cref{eq:costate,eq:supp-costate}) \\
$\lambda_{\text{TD}}$ & $[0,1]$ & $\mathrm{TD}(\lambda)$ eligibility-trace parameter ($\lambda_{\text{TD}}=0.95$; \cref{tab:supp-hyper}) \\
$\lambda_{\text{sol}}$ & $\mathbb{R}^{\text{nv}}$ & Constraint-solver adjoint multiplier $H_{\text{sol}}^{-1}\bar a$ (\cref{eq:supp-ift}) \\
$H$ & $\mathbb{Z}_+$ & Differentiable rollout window horizon ($H=64$ on G1, $H=8$ on DMC) \\
$H_{\text{sol}}$ & $\mathbb{R}^{\text{nv}\times\text{nv}}$ & Newton-solver Hessian $M + \sum_{e\in\mathcal{A}} D_e J_e^\top J_e$ (\cref{eq:supp-hessian}) \\
$d$ & $\mathbb{Z}_+$ & Action dimension ($d=29$ joints on Unitree G1) \\
$N$ & $\mathbb{Z}_+$ & Live window-step samples per batch ($N \le B \times H$) \\
\bottomrule
\end{tabular}}%
\end{table}

\paragraph{Window objective and backward recursion.}
\Cref{alg:supp-nablapo} gives the closed-loop computational graph and \cref{tab:supp-notation} the notation. Let $B$ be the number of parallel window environments and $H$ the window length. At each step $t \in \{0, \dots, H-1\}$, the policy outputs mean action $\mu_t = \pi_\theta(o_t)$ where $o_t = \mathrm{obs}(s_t)$, and the simulator executes $a_t = \mu_t + \epsilon_t$ with $\epsilon_t \sim \mathcal{N}(0, \sigma^2 I_d)$ using the reparameterization trick. The discounted window return is
\begin{equation}
  J = \sum_{t=0}^{H-1} \gamma^t \bigl(r_{\text{task}}(s_t, a_t) + r_{\text{pen}}(s_t)\bigr) + \gamma^H V_\phi(s_H).
  \label{eq:supp-window-return}
\end{equation}
With terminal condition $\lambda_H = \gamma^H \nabla_{s_H} V_\phi(s_H)$, the closed-loop backward recursion for $t = H-1, \dots, 0$ is:
\begin{align}
  g_t &= \frac{\partial J}{\partial a_t} = \gamma^t \frac{\partial r_t}{\partial a_t} + \lambda_{t+1}^{\!\top} \frac{\partial f(s_t, a_t)}{\partial a_t}, \nonumber\\
  \lambda_t &= \gamma^t \frac{\partial r_t}{\partial s_t} + \left(\frac{\partial \pi_\theta(\mathrm{obs}(s_t))}{\partial s_t}\right)^{\!\top} \tilde g_t + \lambda_{t+1}^{\!\top} \frac{\partial f(s_t, a_t)}{\partial s_t}.
  \label{eq:supp-costate}
\end{align}
Three implementation details of \cref{eq:supp-costate}:
\begin{enumerate}[leftmargin=*,itemsep=1pt,topsep=2pt]
  \item \textbf{Clipping before policy feedback:} The per-sample clip (\cref{eq:supp-clip}) is applied to the action cotangent $g_t$ before it is multiplied by $(\partial \pi_\theta(o_t)/\partial s_t)^{\!\top}$ in the costate update.
  \item \textbf{Episode resets and stop-gradient handling:} When an environment falls at step $t$ (pelvis height error $>0.25$\,m or tilt error $>0.8$\,rad), the episode terminates with zero bootstrap value, the environment resets to a new reference frame $s_{\text{reset}}$, and the remainder of the $H$-step window continues collecting transitions for the new episode. Across the reset boundary, $\lambda_{t+1}$ is zeroed ($\mathrm{stop\_grad}$) so gradients never propagate from the post-reset episode into pre-fall actions. To provide a smooth approach gradient before the hard fall threshold is crossed, training adds a quadratic soft termination margin penalty:
  \begin{align}
    r_{\text{pen}}(s_t) = -W\,\Delta t \Biggl[ &\left(\frac{\operatorname{relu}(d_z - d_{z,0})}{h_{\max} - d_{z,0}}\right)^{2} \nonumber\\
      &+ \left(\frac{\operatorname{relu}(d_\theta - d_{\theta,0})}{\theta_{\max} - d_{\theta,0}}\right)^{2} \Biggr]
    \label{eq:supp-term-pen}
  \end{align}
  where $d_z$ and $d_\theta$ are the pelvis height and tilt errors, $W = 2$, $d_{z,0} = 0.08$\,m, $h_{\max}=0.25$\,m, $d_{\theta,0} = 0.10$\,rad, and $\theta_{\max}=0.80$\,rad. Falls enter the gradient only through this smooth margin penalty; the hard termination itself is never differentiated, and the final policies track all 40 LAFAN1 clips without a fall (\cref{tab:supp-lafan-full}).
  \item \textbf{Numerical floor for zero/sparse gradients:} In both in-backward clipping and window-output clipping, we include a numerical floor $\varepsilon = 10^{-8}$,
  \begin{equation}
    \tilde g_i = g_i \min\!\left(1,\ \frac{\kappa\,\operatorname{median}_{j\in\mathcal{L}} \lVert g_j \rVert_2 + \varepsilon}{\lVert g_i \rVert_2 + \varepsilon}\right),
    \label{eq:supp-clip}
  \end{equation}
  where the median is taken over live (non-masked, non-zero) rows $\mathcal{L}$. If all rows in a step have $g_i = 0$ (sparse rewards), then $\tilde g_i = 0$ and $y_t = a_t$.
\end{enumerate}

\subsection{Action-Target Normalization and Step-Size Calibration}
\label{supp:trust-region}

\paragraph{Definition of $\operatorname{RMS}(\tilde g)$ and proof of the $\sqrt{2}\,\kappa\,\delta$ target bound.}
Let $N = |\mathcal{L}|$ be the number of live (step, environment) samples in the window batch ($N \le B \times H = 2048 \times 64 = 131{,}072$ on the humanoid) and $d$ be the action dimension ($d = 29$ joints on the Unitree G1). $\operatorname{RMS}(\tilde g)$ is the root-mean-square over all $N$ samples and all $d$ action coordinates:
\begin{equation}
  \operatorname{RMS}(\tilde g) = \sqrt{\frac{1}{N d} \sum_{i=1}^{N} \lVert \tilde g_i \rVert_2^2 + \varepsilon},
  \qquad
  y_i = a_i + \delta\,\frac{\tilde g_i}{\operatorname{RMS}(\tilde g)}.
  \label{eq:supp-target}
\end{equation}
The batch-wide per-joint RMS target displacement over all $N \times d$ joint targets therefore equals $\delta$ up to $\varepsilon$ (radians on the G1):
\begin{equation}
  \sqrt{\frac{1}{N d} \sum_{i=1}^{N} \lVert y_i - a_i \rVert_2^2} = \delta\,\frac{\sqrt{\frac{1}{Nd}\sum_{i=1}^N \lVert \tilde g_i \rVert_2^2}}{\operatorname{RMS}(\tilde g)} \approx \delta.
\end{equation}
For any individual sample $i \in \{1, \dots, N\}$, its \textbf{per-joint RMS target displacement} is $\Delta_{\text{joint}}(i) \triangleq \frac{1}{\sqrt{d}}\lVert y_i - a_i \rVert_2$. Let $m = \operatorname{median}_{j} \lVert g_j \rVert_2 > 0$. By \cref{eq:supp-clip}, for any $\kappa \ge 1$:
\begin{enumerate}[leftmargin=*,itemsep=1pt,topsep=1pt]
  \item Every sample satisfies $\lVert \tilde g_i \rVert_2 \le \kappa m$.
  \item At least $\lceil N/2 \rceil \ge N/2$ samples have $\lVert g_j \rVert_2 \ge m$, and because $\kappa \ge 1$, their clipped norms also satisfy $\lVert \tilde g_j \rVert_2 = \min(\lVert g_j \rVert_2, \kappa m) \ge m$.
\end{enumerate}
Summing over the $N$ samples gives $\sum_{j=1}^N \lVert \tilde g_j \rVert_2^2 \ge \frac{N}{2} m^2$, which implies $\operatorname{RMS}(\tilde g) \ge \frac{m}{\sqrt{2d}}$. Substituting both inequalities yields the scale-free per-sample, per-joint RMS target bound:
\begin{align}
  \Delta_{\text{joint}}(i) &= \frac{1}{\sqrt{d}}\lVert y_i - a_i \rVert_2 = \frac{\delta\,\lVert \tilde g_i \rVert_2}{\sqrt{d}\,\operatorname{RMS}(\tilde g)} \nonumber\\
  &\le \frac{\delta\,\kappa m}{\sqrt{d}\,(m / \sqrt{2d})} = \sqrt{2}\,\kappa\,\delta.
  \label{eq:supp-bound}
\end{align}
With $\kappa=3$ and $\delta=0.03$\,rad on the G1, every sample's per-joint RMS target displacement is at most $\sqrt{2}\,\kappa\,\delta \approx 0.127$\,rad.

\begin{table*}[t]
\centering
\scriptsize
\caption{\textbf{Deterministic 50-update branch probes from checkpoint $u_0$ across target radii $\{0.5, 1, 2, 4\}\times\delta$ ($\delta=0.03$\,rad).} Evaluated over $4{,}096$ fixed library start frames $\times 250$ deterministic steps ($J_0\in[0,0.1]$ mean step return; noise-free actor). Columns report $\Delta J = J(u_0{+}50) - J_0$ (two branch seeds at $1\times\delta$) and cumulative overshoot cosine $\cos(\text{step}_N, g_N)$ ($>0$: undershoots; $<0$: next gradient points back).}
\label{tab:supp-branch}
\setlength{\tabcolsep}{4pt}
\resizebox{\textwidth}{!}{%
\begin{tabular}{@{}lccccc@{\hspace{10pt}}cccc@{}}
\toprule
& & \multicolumn{4}{c}{50-update deterministic return change $\Delta J = J_{\text{end}} - J_0$} & \multicolumn{4}{c}{Whole-branch overshoot $\cos(\text{step}_N, g_N)$} \\
\cmidrule(lr){3-6}\cmidrule(l){7-10}
Start $u_0$ & $J_0$ & $0.5\times\delta$ & $1\times\delta$ (s0 / s1) & $2\times\delta$ & $4\times\delta$ & $0.5\times\delta$ & $1\times\delta$ (s0 / s1) & $2\times\delta$ & $4\times\delta$ \\
\midrule
$0$ & $0.0083$ & $+0.0144$ & $+0.0244\ /\ +0.0190$ & $+0.0230$ & $\mathbf{+0.0261}$ & $+0.111$ & $+0.034\ /\ +0.067$ & $+0.010$ & $+0.002$ \\
$100$ & $0.0536$ & $+0.0020$ & $+0.0059\ /\ +0.0067$ & $\mathbf{+0.0139}$ & $+0.0129$ & $-0.053$ & $-0.035\ /\ -0.040$ & $-0.020$ & $-0.022$ \\
$250$ & $0.0699$ & $-0.0010$ & $\mathbf{+0.0040}\ /\ +0.0021$ & $\mathbf{+0.0042}$ & $+0.0024$ & $-0.052$ & $-0.033\ /\ -0.045$ & $-0.022$ & $-0.038$ \\
$500$ & $0.0788$ & $-0.0015$ & $+0.0010\ /\ -0.0001$ & $\mathbf{+0.0011}$ & $-0.0001$ & $-0.062$ & $-0.067\ /\ -0.065$ & $-0.059$ & $-0.056$ \\
$1{,}000$ & $0.0866$ & $-0.0013$ & $-0.0003\ /\ -0.0005$ & $-0.0005$ & $-0.0023$ & $-0.066$ & $-0.067\ /\ -0.057$ & $-0.067$ & $-0.089$ \\
$2{,}000$ & $0.0886$ & $-0.0016$ & $-0.0006\ /\ -0.0009$ & $-0.0006$ & $\mathbf{-0.0116}$ & $-0.075$ & $-0.077\ /\ -0.070$ & $-0.076$ & $\mathbf{-0.144}$ \\
$4{,}000$ & $0.0891$ & $-0.0010$ & $-0.0003\ /\ -0.0001$ & $-0.0009$ & $-0.0079$ & $-0.074$ & $-0.072\ /\ -0.069$ & $-0.092$ & $\mathbf{-0.139}$ \\
$12{,}000$ & $0.0918$ & $-0.0001$ & $-0.0001\ /\ -0.0001$ & $-0.0001$ & $-0.0023$ & $-0.043$ & $-0.048\ /\ -0.034$ & $-0.053$ & $\mathbf{-0.140}$ \\
\midrule
\multicolumn{10}{@{}l}{\emph{Late-stage collapse recovery (one run, resumed from update $18{,}000$, full-clip return $0.886$):}} \\
\multicolumn{5}{@{}l}{Full regression step ($18\text{k}\to 20\text{k}\to 22\text{k}\to 24\text{k}\to 25\text{k}$): $0.886 \to 0.858 \to 0.832 \to 0.744 \to 0.653$} & \multicolumn{5}{l@{}}{Regression step $\times 0.05$, $\delta$ unchanged ($18\text{k}\to 20\text{k}\to 22\text{k}\to 24\text{k}\text{--}25\text{k}$): $0.886 \to 0.907 \to 0.914 \to \mathbf{0.915}$ (0 falls)} \\
\bottomrule
\end{tabular}}%
\end{table*}

\paragraph{Deterministic branch-probe calibration of $\delta$ (\cref{tab:supp-branch}).}
Because \engine is bitwise deterministic, we launch 50-update branches from eight checkpoints ($u_0 \in \{0, \dots, 12{,}000\}$) with identical start states and RNG seeds. At initialization ($u_0=0$), $4\times\delta$ produces the largest gain ($+0.0261$) and all arms undershoot ($\cos > 0$). At $u_0=100$, $2\times$--$4\times\delta$ still double the progress of $1\times\delta$ ($+0.0139$ vs.\ $+0.0059$--$0.0067$), whereas $1\times$ and $2\times$ tie by $u_0=250$ ($+0.0040$ vs.\ $+0.0042$). From $u_0\ge 1{,}000$ onward, $4\times\delta$ overshoots the narrower tracking basin ($\Delta J = -0.0023$ to $-0.0116$, overshoot $\cos = -0.089$ to $-0.144$). In the late-collapse run (bottom rows), resuming a pre-collapse checkpoint ($0.886$) with the full regression step drops steadily ($0.858 \to 0.832 \to 0.653$), whereas scaling the regression optimizer's parameter update by $0.05$ with $\delta$ unchanged climbs monotonically to $0.915$ ($0$ falls). Both probes call for smaller steps later in training; we therefore decay $\delta$ from $0.03$ to $0.009$\,rad together with the learning-rate schedule (\cref{tab:supp-hyper}). In the final recipe, whose regression learning rate decays as well, holding $\delta$ fixed over 24{,}000 updates reaches the same plateau ($0.932\pm0.001$ vs.\ $0.932\pm0.002$, 3 seeds).

\section{Component Ablations}
\label{supp:ablations}

\subsection{Exact-Adjoint Component Ablations on 40-Clip LAFAN1}
\label{supp:exact-ablations}

\begin{figure*}[t]
  \centering
  \includegraphics[width=0.92\linewidth]{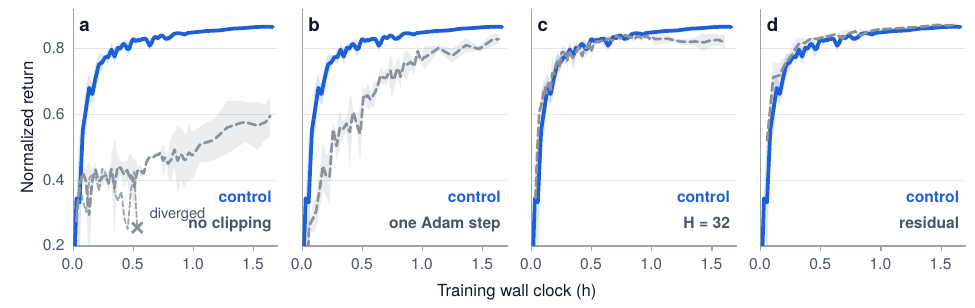}
  \caption{\textbf{Component ablations of \method on exact \engine adjoints (40-clip LAFAN1, absolute actions, 3 seeds, one RTX PRO 6000).} Control in blue (mean $\pm$ s.d.), one-component variant in grey dashed: (a)~global norm clipping instead of per-sample clipping; (b)~one Adam step per window instead of $M=8$ regression steps; (c)~$H=32$ instead of $H=64$; (d)~residual instead of absolute actions. Numbers are in \cref{tab:supp-abl-abs}.}
  \label{fig:supp-ablations}
\end{figure*}

\begin{table*}[t]
\centering
\scriptsize
\caption{\textbf{Exact-adjoint component ablations on the earlier residual-action recipe (40-clip LAFAN1, 3 seeds per arm, $\sim 1.6$\,h on one RTX PRO 6000; control $0.871$).} Each row changes one component; everything else is held fixed.}
\label{tab:supp-abl-res}
\setlength{\tabcolsep}{4.5pt}
\resizebox{\textwidth}{!}{%
\begin{tabular}{@{}llcccccc@{}}
\toprule
Ablated component (single change from control) & Origin & Seeds & Matched at & Control return & Variant return & $\Delta$ return & Failed clips (/40): ctrl $\to$ var \\
\midrule
No per-sample clipping (global norm clip only) & Ours & 0 of 3 survived\textsuperscript{a} & --- & --- & Diverged (3/3) & --- & --- \\
No multi-step target regression (1 Adam step/window) & Ours & 3 & 1.63\,h & $0.871\pm 0.001$ & $0.830\pm 0.015$ & $\mathbf{-0.041}$ & $0.1\pm 0.1 \to 1.3\pm 0.8$ \\
$H=32$ window instead of $H=64$ (matched wall clock) & Ours & 3 of 4\textsuperscript{b} & 1.65\,h & $0.871\pm 0.001$ & $0.831\pm 0.009$ & $\mathbf{-0.040}$ & $0.1\pm 0.1 \to 0.3\pm 0.2$ \\
$H=32$ window instead of $H=64$ (matched updates: 25.6k) & Ours & 3 of 4\textsuperscript{b} & 25.6k upd. & $0.878\pm 0.003$ & $0.831\pm 0.009$ & $\mathbf{-0.046}$ & $0.0\pm 0.0 \to 0.3\pm 0.2$ \\
\bottomrule
\end{tabular}}%
\par\smallskip\raggedright\scriptsize
\textsuperscript{a}\,All three seeds diverged (non-finite loss) within 0.31--0.59\,h. \quad
\textsuperscript{b}\,A fourth seed collapsed (critic value runaway) and is excluded.
\end{table*}

\begin{table*}[t]
\centering
\scriptsize
\caption{\textbf{Exact-adjoint component ablations on the earlier absolute-action recipe (40-clip LAFAN1, 3 seeds per arm, $\sim 1.6$\,h on one RTX PRO 6000; curves in \cref{fig:supp-ablations}).}}
\label{tab:supp-abl-abs}
\setlength{\tabcolsep}{4.5pt}
\resizebox{\textwidth}{!}{%
\begin{tabular}{@{}llcccccc@{}}
\toprule
Ablated component (single change from control) & Origin & Seeds & Matched at & Control return & Variant return & $\Delta$ return & Failed clips (/40): ctrl $\to$ var \\
\midrule
No per-sample clipping (global norm clip only) & Ours & 2 of 3\textsuperscript{a} & 1.65\,h & $0.862\pm 0.003$ & $0.570\pm 0.077$ & $\mathbf{-0.292}$ & $0.0\pm 0.0 \to 10.4\pm 5.1$ \\
No multi-step target regression (1 Adam step/window) & Ours & 3 & 1.64\,h & $0.862\pm 0.002$ & $0.809\pm 0.010$ & $\mathbf{-0.053}$ & $0.0\pm 0.0 \to 2.5\pm 1.0$ \\
$H=32$ window instead of $H=64$ (matched wall clock) & Ours & 3 & 1.64\,h & $0.862\pm 0.002$ & $0.821\pm 0.019$ & $\mathbf{-0.040}$ & $0.0\pm 0.0 \to 0.4\pm 0.2$ \\
$H=32$ window instead of $H=64$ (matched updates: 26k) & Ours & 3 & 26.0k upd. & $0.868\pm 0.002$ & $0.821\pm 0.019$ & $\mathbf{-0.046}$ & $0.1\pm 0.2 \to 0.4\pm 0.2$ \\
Residual actions instead of absolute setpoints & Param. & 3 & 1.64\,h & $0.862\pm 0.002$ & $0.870\pm 0.001$ & $+0.008$ & $0.0\pm 0.0 \to 0.1\pm 0.1$ \\
\bottomrule
\end{tabular}}%
\par\smallskip\raggedright\scriptsize
\textsuperscript{a}\,Seed 0 diverged at 0.54\,h.
\end{table*}

\begin{table}[t]
\centering
\scriptsize
\caption{\textbf{Sensitivity and mechanism ablations on exact \engine adjoints (earlier residual recipe, 1.5\,h on one RTX PRO 6000, mean $\pm$ s.d.\ over 3 seeds).} Default recipe values in bold.}
\label{tab:supp-sensitivity}
\setlength{\tabcolsep}{3.5pt}
\resizebox{\linewidth}{!}{%
\begin{tabular}{@{}llll@{}}
\toprule
Factor & Setting & 1.5\,h Return & Observation / crossing time \\
\midrule
Clipping threshold $\kappa$ & $\kappa = 1$ & $0.853\pm 0.008$ & Stable (all 3 seeds) \\
 & $\mathbf{\kappa = 3}$ & $\mathbf{0.869\pm 0.003}$ & Default (all tasks) \\
 & $\kappa = 10$ & $0.849\pm 0.011$ & Stable (all 3 seeds) \\
 & $\kappa = 30$ & $0.849\pm 0.005$ & Stable (all 3 seeds) \\
 & None (global clip) & Diverged & 3 of 3 seeds diverged ($<0.6$\,h) \\
\midrule
Engine gradient rules & \textbf{\engine training mode} & $\mathbf{0.869\pm 0.003}$ & Two geometric rules (\cref{supp:engine-accuracy}) \\
 & Unmodified exact & $0.867\pm 0.002$ & Same time to 0.82 and update rate \\
\midrule
Window horizon $H$ & $H = 16$ & Diverged & 3 of 3 diverged (critic value runaway) \\
 & $H = 32$ & $0.815$--$0.841$ & 1 of 4 collapsed (critic runaway) \\
 & $\mathbf{H = 64}$ & $\mathbf{0.869\pm 0.003}$ & 0 collapses ($\sim$14.3k updates) \\
 & $H = 128$ & $0.839\pm 0.004$ & Stable ($\sim 7.9\text{k}$ updates, $0.55\times$ update throughput) \\
\midrule
Windows per update $B$ & $B = 1024$ & $0.865\pm 0.002$ & $\sim 27.0$k updates in 1.5\,h \\
 & $B = 2048$ & $0.867\pm 0.004$ & $\sim 20.3$k updates in 1.5\,h \\
 & $\mathbf{B = 4096}$ & $\mathbf{0.869\pm 0.003}$ & $\sim 14.3$k updates in 1.5\,h \\
 & $B = 8192$ & $0.852\pm 0.002$ & $\sim 8.6$k updates in 1.5\,h \\
\bottomrule
\end{tabular}}%
\end{table}

\begin{figure}[t]
  \centering
  \includegraphics[width=0.98\linewidth]{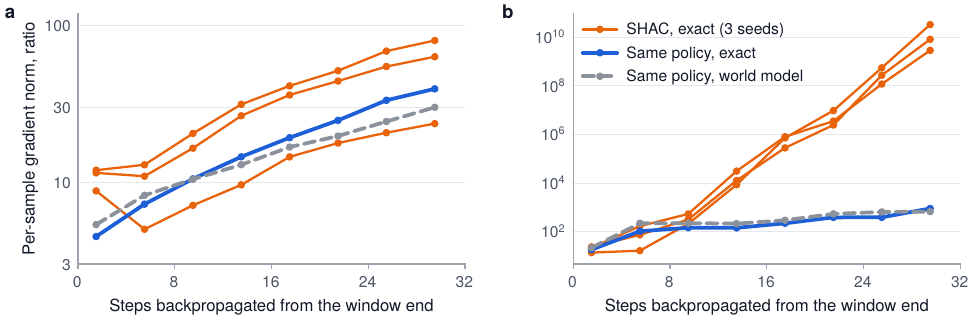}
  \caption{\textbf{Per-sample action-gradient tails relative to the batch median, grow exponentially with backpropagation depth $k\in[1,32]$.} (a)~99.9th percentile and (b)~maximum norm relative to the batch median as a function of steps backpropagated from the window end: raw exact gradients in SHAC (orange, 3 seeds) grow by $10^8$--$10^{10}\times$ over 32 steps; on a converged policy, exact (\engine, blue) and learned world-model (grey dashed) gradients show the same tail growth with depth.}
  \label{fig:supp-tails-depth}
\end{figure}

\begin{table}[t]
\centering
\scriptsize
\caption{\textbf{Component ablations of the final recipe on 40-clip LAFAN1 (3 seeds per arm, one A40 GPU, about 54\,min of training each).} Hold: first evaluation after which every later evaluation stays at or above the level (evaluations every 200 updates, MuJoCo Warp). Returns and falls are seed means. Without per-sample clipping, the regression optimizer keeps its global-norm clip (1.0). Fixed scale: the action-target step uses a constant gain equal to the control run's early median instead of $\delta/\mathrm{RMS}$, with the same $\delta$ schedule.}
\label{tab:supp-abl-final}
\setlength{\tabcolsep}{3pt}
\resizebox{\linewidth}{!}{%
\begin{tabular}{@{}lccccc@{}}
\toprule
Arm & Hold 0.866 (min) & Hold 0.890 (min) & Return, 30\,min & Return, 54\,min & Falls /40 \\
\midrule
\textbf{Full recipe} & $\mathbf{13.9\pm 1.3}$ & $\mathbf{38.7\pm 3.3}$ (3/3) & $\mathbf{0.887}$ & $\mathbf{0.897}$ & 0.3 \\
No per-sample clipping & never & never & 0.034 & 0.058 & 39.7 \\
Window-output clipping only & $13.1\pm 1.4$ & $39.3\pm 3.8$ (3/3) & 0.884 & 0.900 & 0 \\
Single regression step ($M=1$) & $21.4\pm 0.8$ & $51.7\pm 2.1$ (2/3) & 0.877 & 0.889 & 0 \\
No RMS normalization (fixed scale) & $20.0\pm 1.3$ & $43.4$ (1/3) & 0.878 & 0.888 & 0 \\
\bottomrule
\end{tabular}}%
\end{table}

\Cref{tab:supp-abl-final} ablates the core components of the final recipe. Without per-sample clipping, every seed collapses within about 100 updates: a few rows with very large gradients dominate the batch RMS, so the normalized step $\delta/\mathrm{RMS}$ shrinks toward zero for all other samples. Applying the clip only to the window's pooled action gradients (step~4 of \cref{alg:supp-nablapo}) instead of also at every backward step matches the full recipe, and does so again over a full 24{,}000-update run (3 seeds, RTX PRO 6000: plateau $0.930\pm0.003$ vs.\ $0.932\pm0.002$, 0 falls); the component is the per-sample, median-relative threshold, and the in-backward placement is how our implementation applies it. A single regression step per update ($M=1$) and a fixed step scale without RMS normalization both train stably but reach $0.866$ about 6--8\,min later and $0.890$ on fewer seeds. We also ablated each ingredient with 3 seeds on two earlier recipes on 40-clip LAFAN1 in \engine (\cref{tab:supp-abl-res,tab:supp-abl-abs,tab:supp-sensitivity} and \cref{fig:supp-ablations}):
\begin{enumerate}[leftmargin=*,itemsep=1.5pt,topsep=2pt]
  \item \textbf{Per-sample clipping vs.\ global norm clipping:} Removing per-sample clipping (retaining global gradient-norm clipping) causes all 3 residual-action seeds to diverge within $0.31$--$0.59$\,h (\cref{tab:supp-abl-res}), and causes 1 of 3 absolute-action seeds to diverge while the remaining two stall at $0.570\pm 0.077$ vs.\ $0.862\pm 0.003$ (\cref{tab:supp-abl-abs}, \cref{fig:supp-ablations}a). Every tested threshold $\kappa \in \{1, 10, 30\}$ also prevents divergence and finishes $0.016$--$0.020$ below $\kappa=3$ (\cref{tab:supp-sensitivity}). Without clipping, per-sample tails grow by $10^8$--$10^{10}\times$ over 32 backpropagated steps (\cref{fig:supp-tails-depth}).
  \item \textbf{Multi-step action-target regression ($M=8$ with RMS normalization) vs.\ single Adam step ($M=1$):} With $M=1$ minibatch step and matched learning rate, action-target regression reduces to a single direct gradient step per window~\cite{mora2021pods,you2025dva} (\ie SHAC equipped with our per-sample clipping and $H=64$ window). Replacing $M=8$ regression steps with 1 Adam step per window lowers the 1.6\,h return by $0.041$ on residual actions ($0.871 \to 0.830$, failed clips increasing from $0.1$ to $1.3$; \cref{tab:supp-abl-res}) and by $0.053$ on absolute actions ($0.862 \to 0.809$, failed clips $0.0 \to 2.5$; \cref{tab:supp-abl-abs}, \cref{fig:supp-ablations}b).
  \item \textbf{Window horizon $H$:} Short windows rely more on the critic's value gradient $\nabla_{s_H} V_\phi(s_H)$: at $H=16$, all 3 seeds diverge via critic value runaway; at $H=32$, 1 of 4 seeds collapses and the surviving seeds lose $0.040$ at matched wall clock ($0.046$ at matched updates; \cref{tab:supp-abl-res,tab:supp-sensitivity}). At $H=128$, training is stable but reaches only $0.839\pm 0.004$, since each update costs roughly twice as much simulation ($0.55\times$ throughput, $\sim 7.9\text{k}$ updates in the same budget). $H=64$ balances stability and throughput.
  \item \textbf{Geometric rules:} Training \method with the exact-mode adjoint (both geometric rules of \cref{supp:engine-accuracy} off) gives $0.867\pm 0.002$ vs.\ $0.869\pm 0.003$ after 1.5\,h across 3 seeds, with the same time to reach $0.82$ and no diverged seed (\cref{tab:supp-sensitivity}).
\end{enumerate}

\subsection{Horizon and Clipping with World-Model Gradients}
\label{supp:wm-decoupling}

\begin{table}[t]
\centering
\scriptsize
\caption{\textbf{Decoupling window horizon ($H=32$ vs.\ $H=64$) and per-sample clipping ($\kappa=3$) on world-model gradients (\napo-WM vs.\ InstantTrack v1.4, 3 seeds per arm, one A40).} Time in minutes to reach each return threshold (with 0 failed clips) and return at 90\,min.}
\label{tab:supp-wm-ablation}
\setlength{\tabcolsep}{3.2pt}
\resizebox{\linewidth}{!}{%
\begin{tabular}{@{}lccccccc@{}}
\toprule
Method (A40 GPU, 3 seeds) & $H$ & Clip $\kappa=3$ & s/upd. & To 0.82 & To 0.84 & To 0.86 & 90\,min Return \\
\midrule
InstantTrack v1.4 (official) & 32 & $\times$ & 0.17\,s & \textbf{9.0} & 22.0 & 68.0 & $0.859\pm 0.002$ \\
\napo-WM ($H=32$, clip only) & 32 & \checkmark & 0.25\,s & 10.6 & \textbf{11.4} & 50.9 & $0.868\pm 0.003$ \\
\napo-WM ($H=64$ + clip) & 64 & \checkmark & 0.44\,s & 21.7 & 21.7 & \textbf{24.6} & $\mathbf{0.883\pm 0.002}$ \\
\bottomrule
\end{tabular}}%
\end{table}

\begin{table}[t]
\centering
\scriptsize
\caption{\textbf{Directional cosine similarity between surrogate/world-model action gradients and \engine's exact action gradient $\partial J/\partial a_t$ across tasks and training stages.}}
\label{tab:supp-cosine}
\setlength{\tabcolsep}{3.5pt}
\resizebox{\linewidth}{!}{%
\begin{tabular}{@{}llccc@{}}
\toprule
Gradient source & Task / trajectory & 1k upd. & 4k upd. & 12k upd. \\
\midrule
\engine training mode vs.\ exact mode & LAFAN1 ($H=32$, 65k rows) & \multicolumn{3}{c}{Median cosine $1.0000$ (Spearman $0.9999$)} \\
Smoothed height-scan VJP ($\sigma=1$ cell) & Terrain (320 poses, interior) & \multicolumn{3}{c}{Cosine $0.985$ (unsmoothed: $0.26$--$0.36$)} \\
Strong WM (\napo window $H=64$) & Velocity tracking (online) & $0.81$ & $0.93$ & $0.94$ ($5\%$ opp.\ sign) \\
Strong WM (InstantTrack window $H=32$) & Velocity tracking (online) & $0.64$ & $0.87$ & $0.91$ ($13\%$ opp.\ sign) \\
Weak WM ($3\times 2048$ MLP, no LayerNorm) & Online & \multicolumn{3}{c}{$0.08$ early; $0.40$--$0.47$ later ($13$--$30\%$ opp.)} \\
\bottomrule
\end{tabular}}%
\end{table}

SuperTrack~\cite{fussell2021supertrack}, SuperMimic~\cite{liao2026supermimic} and InstantTrack~\cite{dao2026instanttrack} optimize a tracking policy through a learned differentiable model of the simulator; \napo-WM uses the same kind of gradient inside the \napo update. \Cref{fig:axis}c compares \napo-WM ($H=64$, per-sample clipping) with an unclipped $H=32$ variant that matches InstantTrack v1.4's window. \Cref{tab:supp-wm-ablation} adds the two changes one at a time:
\begin{itemize}[leftmargin=*,itemsep=1pt,topsep=2pt]
  \item At $H=32$, adding \napo's per-sample clipping ($\kappa=3$) alone nearly halves the time to reach $0.84$ (from $22.0$ to $11.4$\,min), cuts the time to reach $0.86$ from $68.0$\,min to $50.9$\,min, and raises the 90-minute return from $0.859$ to $0.868$.
  \item Extending the clipped window from $H=32$ to $H=64$ further cuts the time to reach $0.86$ from $50.9$\,min to $24.6$\,min ($2.8\times$ faster than InstantTrack v1.4 on the same GPU) and raises the 90-minute return to $0.883$.
  \item At matched policy updates on LAFAN1 (4{,}000 updates; 3 seeds per world-model arm, 5 for the exact adjoint), InstantTrack settings ($H=32$, no clip) reach $0.828$, adding the $H=64$ window reaches $0.853$, and adding per-sample clipping ($\kappa=3$) reaches $0.869$; with \engine's exact adjoints, \napo reaches $0.883$ at 4{,}000 updates and $0.914$ at 12{,}000 updates.
\end{itemize}
Gradient cosines against the exact adjoint are listed in \cref{tab:supp-cosine}. Per policy update, the exact adjoint gives the larger improvement; a world-model update is cheaper because the learned model replaces the differentiable simulator step, which is where its wall-clock advantage comes from.

\subsection{Per-Update Cost on Two Engines}
\label{supp:sample-efficiency}

\begin{figure}[t]
  \centering
  \includegraphics[width=0.98\linewidth]{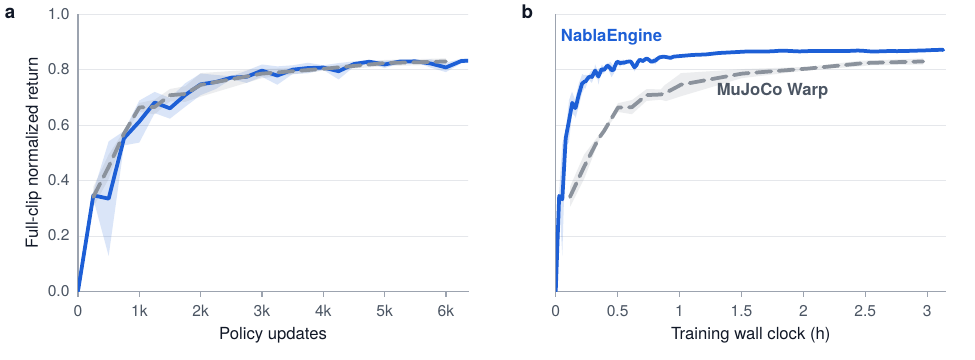}
  \caption{\textbf{Same \method algorithm trained on two differentiable simulators: \engine (blue) vs.\ MuJoCo Warp adjoint (grey dashed) across 3 seeds on one RTX PRO 6000.} (a)~Against policy updates, the two learning curves overlap throughout training (3-seed means differ by $\le 0.02$). (b)~Against wall-clock training hours, \engine completes 4{,}000 updates in $0.43$\,h vs.\ $2.02$\,h ($4.7\times$ faster).}
  \label{fig:supp-operator}
\end{figure}

\begin{table*}[t]
\centering
\scriptsize
\caption{\textbf{End-to-end training-loop cost per policy update across methods, physics operators, and GPUs.} MJW: MuJoCo Warp.}
\label{tab:supp-cost}
\setlength{\tabcolsep}{4.2pt}
\resizebox{\textwidth}{!}{%
\begin{tabular}{@{}lllcccccc@{}}
\toprule
Method & Physics operator & GPU & Median s/update & Updates / hour (measured) & Diff.\ steps / upd. & Fwd-only steps / upd. & Total sim.\ steps / s & Speedup vs.\ MJW \\
\midrule
\method (residual, $B=2048$) & \engine & RTX PRO 6000 & \textbf{0.240\,s} & 14{,}384 & 131{,}072 & 10{,}240 & 588k & --- \\
\method (absolute, $B=4096$) & \engine & RTX PRO 6000 & \textbf{0.373\,s} & 9{,}383 & 262{,}144 & 10{,}240 & 729k & $\mathbf{4.4\times}$ \\
\method (absolute, $B=4096$) & MuJoCo Warp & RTX PRO 6000 & 1.633\,s & 1{,}966 & 262{,}144 & 10{,}240 & 167k & $1.0\times$ \\
\midrule
\method (absolute, $B=4096$) & \engine & H100 & \textbf{0.382\,s} & 9{,}275 & 262{,}144 & 10{,}240 & 712k & $\mathbf{5.9\times}$ \\
\method (absolute, $B=4096$) & MuJoCo Warp & H100 & 2.258\,s & 1{,}594 & 262{,}144 & 10{,}240 & 121k & $1.0\times$ \\
\midrule
SHAC ($B=4096, H=32$) & \engine & RTX PRO 6000 & 0.196\,s & 18{,}365 & 131{,}072 & 0 & 669k & $4.3\times$ \\
SHAC ($B=4096, H=32$) & MuJoCo Warp & RTX PRO 6000 & 0.839\,s & 3{,}455 & 131{,}072 & 0 & 156k & $1.0\times$ \\
\midrule
PPO (BeyondMimic, $4096\times 24$) & \engine (fwd) & RTX PRO 6000 & 0.126\,s/iter & 28{,}240 & 0 & 98{,}304 & 781k & $2.9\times$ \\
PPO (SuperMimic actor, $8192\times 24$) & \engine (fwd) & RTX PRO 6000 & 0.549\,s/iter & 6{,}542 & 0 & 196{,}608 & 358k & --- \\
\bottomrule
\end{tabular}}%
\end{table*}

Running \method (absolute actions) on the MuJoCo Warp adjoint (PR \#1535 with follow-up fixes~\cite{mjwarp_adjointfix2026}) gives nearly the same return per policy update (4{,}548 updates on \engine vs.\ 4{,}802 on MuJoCo Warp to reach $0.82$; \cref{fig:supp-operator}), while \engine reduces the per-update wall-clock cost from $1.633$\,s to $0.373$\,s ($4.4\times$ per update, $4.7\times$ in wall-clock time to 4{,}000 updates; \cref{tab:supp-cost}). The final residual recipe ($B=2048$) takes $0.240$\,s per update on \engine.

\FloatBarrier
\section{\engine Architecture, Implicit-Function-Theorem Adjoints, and Verification}
\label{supp:engine}

\subsection{Model, Kernel Architecture, and Constraint-Solve Adjoint Derivation}
\label{supp:engine-derivation}

\paragraph{Simulated model and MuJoCo physics.}
\engine generates CUDA source from each MuJoCo model, compiles it once and caches it per GPU architecture; no task has hand-written kernels. A tree build covers single-tree models (free root plus hinges) and is used for the G1; a forest build covers free, hinge and slide roots and multiple trees and is used for the DeepMind Control and Brax models and the G1 with a box. Supported geoms are plane, sphere, capsule, box, cylinder and convex mesh, with pyramidal or elliptic cones (condim 1, 3, 4) and the \texttt{implicitfast} (both builds) or Euler (forest) integrator. For the experiments in this paper, \engine simulates the mjlab Unitree G1 model~\cite{zakka2026mjlab}: a floating base and 29 actuated hinge joints ($\text{nq}=36, \text{nv}=35, d=29$), 31 rigid bodies, and 502 contact pairs among capsules, spheres, and the ground plane, with full robot self-collision enabled. The forward dynamics implement MuJoCo's equations of motion~\cite{todorov2012mujoco}:
(i)~recursive Newton--Euler (RNE) and composite rigid-body algorithm (CRBA) kinematics and dynamics;
(ii)~MuJoCo's soft-constraint contact model (\texttt{solref}/\texttt{solimp}) with pyramidal friction cones;
(iii)~joint limits and PD position actuators with force limits (the G1 model has no friction-loss rows);
(iv)~Newton's method with exact line search, stopping when the cost decrease or the gradient norm, both scaled by $1/(\bar m\,\text{nv})$ with mean inertia $\bar m$, falls below $10^{-8}$, or after 10 iterations (on average $4.0$--$4.5$ iterations per substep); and
(v)~the \texttt{implicitfast} integrator with 4 substeps of $5$\,ms per $20$\,ms control step (control held across the 4 substeps). Per-world foot friction and torso center-of-mass offsets are stored per environment for domain randomization (they are not differentiated).

\paragraph{Warp-per-world GPU kernel layout and tape.}
Each parallel world maps to one CUDA thread block of 32 threads (a single warp). Within a kernel, the world's state resides in registers and shared memory ($6.9$\,kB and $8.2$\,kB per world for the two forward kernels), while model constants reside in constant memory. To maximize SM occupancy (11--12 worlds resident per SM on RTX PRO 6000 and H100), each $5$\,ms substep executes two forward kernels: the first computes kinematics, bias and mass matrices, actuator forces, collision detection, and constraint rows; the second runs the 10-iteration Newton solve and \texttt{implicitfast} integration. During differentiated windows, the forward pass writes a compact tape of $1{,}472$ float32 values per world per substep ($23.6$\,kB per world per $20$\,ms control step, or $96.5$\,MB total across 4096 worlds; $6.17$\,GB for an $H=64$ window). The backward pass of an entire $20$\,ms control step (all 4 substeps in reverse) executes as a single fused megakernel: one warp per world reads the tape and differentiates integration, the constraint solve, collision geometry, and kinematics/dynamics in shared memory without global memory round-trips. All reductions use deterministic fixed-order warp shuffles without floating-point atomics, so forward outputs and backward gradients are bitwise identical across repeated runs and processes.

\paragraph{Implicit-function-theorem (IFT) adjoint of the Newton contact solver.}
Per substep, with state $(q, v)$ and control $u$, let $M(q) \in \mathbb{R}^{\text{nv}\times\text{nv}}$ be the joint-space mass matrix, $c(q,v)$ the bias force, $\tau(q,v,u)$ the clamped PD actuator force, and $a_0 = M^{-1}(\tau - c)$ the unconstrained acceleration. MuJoCo's Newton solver minimizes the strictly convex unconstrained objective over joint accelerations $a \in \mathbb{R}^{\text{nv}}$:
\begin{equation}
  a^\star = \arg\min_a \frac{1}{2} (a - a_0)^\top M (a - a_0) + \sum_{e} s_e(J_e a - r_e),
  \label{eq:supp-opt}
\end{equation}
over constraint rows $e$ (joint limits and the edges of the pyramidal contact friction cones), with constraint Jacobians $J_e(q)$, reference accelerations $r_e(q,v) = -b_e J_e v - k_e \iota_e d_e$, and quadratic penalty $s_e(x) = \frac{1}{2} D_e x^2$ on the active side $e \in \mathcal{A}$ of row $e$ ($0$ otherwise, with impedance $D_e(d_e)$). Rather than unrolling the 10 Newton iterations and line searches on a tape, \engine differentiates the first-order optimality condition at the converged acceleration $a^\star$ with the active constraint set $\mathcal{A}$ held fixed:
\begin{equation}
  F(a^\star, \vartheta) = M(a^\star - a_0) + \sum_{e\in\mathcal{A}} D_e J_e^\top (J_e a^\star - r_e) = 0,
  \label{eq:supp-hessian}
\end{equation}
where $\vartheta = (q, v, u)$.
Because $M \succ 0$ and $D_e > 0$, the solver Hessian $H_{\text{sol}} = \frac{\partial F}{\partial a} = M + \sum_{e\in\mathcal{A}} D_e J_e^\top J_e$ is symmetric positive definite and is already Cholesky-factored at the end of the forward Newton solve. If the last Newton step changed the active set of any row, the forward pass refactors once at the final active set without taking a step ($0.10$--$0.17\%$ of records), so the stored factor always matches the active set used by the adjoint. By the implicit function theorem, $\frac{\mathrm{d}a^\star}{\mathrm{d}\vartheta} = -H_{\text{sol}}^{-1} \frac{\partial F}{\partial \vartheta}$. Given an incoming acceleration cotangent $\bar a = \partial J / \partial a^\star$, the adjoint solves a single linear system with the stored Cholesky factor of $H_{\text{sol}}$:
\begin{equation}
  \lambda_{\text{sol}} = H_{\text{sol}}^{-1} \bar a, \quad
  \bar\vartheta = -\nabla_\vartheta \bigl[ \lambda_{\text{sol}}^\top F(a^\star, \vartheta) \bigr],
  \label{eq:supp-ift}
\end{equation}
with $a^\star$ and $\lambda_{\text{sol}}$ held constant.
Substituting $M a_0 = \tau(q,v,u) - c(q,v)$ and inverse dynamics $\mathrm{ID}(q,v,a^\star) = M(q)a^\star + c(q,v)$ into $\lambda_{\text{sol}}^\top F$ yields:
\begin{align}
  \lambda_{\text{sol}}^\top F(a^\star, \vartheta) &= \lambda_{\text{sol}}^\top \mathrm{ID}(q, v, a^\star) - \lambda_{\text{sol}}^\top \tau(q, v, u) \nonumber\\
  &\quad + \sum_{e\in\mathcal{A}} D_e (J_e \lambda_{\text{sol}})(J_e a^\star - r_e).
  \label{eq:supp-id}
\end{align}
Thus $\bar\vartheta = (\bar q, \bar v, \bar u)$ is evaluated in one reverse pass as the sum of three terms:
(1)~the gradient of $\lambda_{\text{sol}}^\top \mathrm{ID}(q,v,a^\star)$, computed by one reverse pass of recursive Newton--Euler;
(2)~the gradient of $-\lambda_{\text{sol}}^\top \tau(q,v,u)$; and
(3)~the gradient of the active constraint rows through $(J_e(q), d_e(q), D_e(d_e))$ and the collision geometry (contact point, normal, and penetration distance).
Finally, the \texttt{implicitfast} step ($Q a_{\text{adv}} = M a^\star$ with $Q = M + h K_v$, $v' = v + h a_{\text{adv}}$, $q' = q \oplus h v'$) is differentiated analytically via $y = Q^{-1}\bar a_{\text{adv}}$ and $\bar a^\star = M y$ plus $\nabla_q[y^\top M(q)(a^\star - a_{\text{adv}})]$.

\subsection{Throughput Benchmarks and Forward/Backward Timing Split}
\label{supp:engine-speed}

\begin{table}[t]
  \centering
  \scriptsize
  \caption{\textbf{Simulator time per 20\,ms control step for 4096 G1 worlds (ms; 10 Newton iterations, 4 substeps).} \engine and MuJoCo Warp: mean of the medians on two held-out state sets, as in \cref{fig:engine}, except V100 (one set). Other simulators: one held-out state set. ---: not measured.}
  \label{tab:supp-engine-fwd}
  \setlength{\tabcolsep}{2.6pt}
\resizebox{\linewidth}{!}{%
  \begin{tabular}{@{}lrrrrrr@{}}
    \toprule
    Forward step & RTX 5090 & RTX PRO 6000 & B300 & H100 & RTX 3090 & V100 \\
    \midrule
    \textbf{\engine (ours)} & \textbf{1.12} & \textbf{1.23} & \textbf{1.47} & \textbf{1.72} & \textbf{3.42} & \textbf{5.79} \\
    MuJoCo Warp 3.17.0 & 5.46 & 6.17 & 7.00 & 8.69 & 12.88 & 18.4 \\
    Genesis 1.4.5 & --- & 13.45 & --- & 16.36 & 36.73 & 139.11 \\
    MJX 3.17 (same 502-pair model) & --- & 255.0 & --- & 216.0 & 510.3 & 710.4 \\
    MJX 3.17 (Menagerie G1, fewer pairs) & --- & 50.7 & --- & 58.4 & 129.2 & 150.7 \\
    Isaac Lab 3.0, Newton (3 box colliders) & --- & 12.9 & --- & --- & 35.2 & --- \\
    Isaac Lab 3.0, PhysX (3 box colliders) & --- & 27.5 & --- & --- & 45.2 & --- \\
    Isaac Lab 3.0, PhysX (38 shapes, self-coll.) & --- & 54.1 & --- & --- & 87.5 & --- \\
    MuJoCo C 3.17 (CPU float64, 16-thread EPYC) & \multicolumn{6}{c}{34.7 (CPU)} \\
    \midrule
    Gradient of one step (fwd with tape + adjoint) & & & & & & \\
    \midrule
    \textbf{\engine (ours)} & \textbf{1.64} & \textbf{1.78} & \textbf{2.34} & \textbf{2.74} & \textbf{5.26} & \textbf{10.6} \\
    MuJoCo Warp + PR \#1535~\cite{mjwarp_pr1535} & 18.96 & 20.98 & 19.25 & 25.13 & 42.74 & 63.2 \\
    Speedup & $\mathbf{11.6\times}$ & $\mathbf{11.8\times}$ & $\mathbf{8.2\times}$ & $\mathbf{9.2\times}$ & $\mathbf{8.1\times}$ & $\mathbf{6.0\times}$ \\
    \bottomrule
  \end{tabular}}%
\end{table}

\begin{figure}[t]
  \centering
  \includegraphics[width=0.98\linewidth]{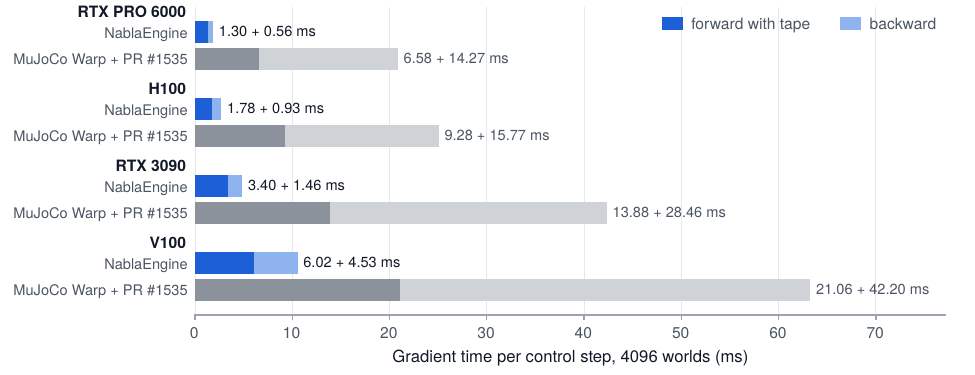}
  \caption{\textbf{Breakdown of 1-step gradient time into forward-with-tape (dark) and backward adjoint (light) kernel calls across 4096 G1 worlds.} On RTX PRO 6000 the two calls take $1.30 + 0.56$\,ms when measured as separate calls; the fused graph in \cref{tab:supp-engine-fwd} takes $1.78$\,ms.}
  \label{fig:supp-engine-split}
\end{figure}

\Cref{tab:supp-engine-fwd} and \cref{fig:supp-engine-split} compare \engine against existing GPU simulators. \engine, MuJoCo Warp and MJX (502-pair rows) use the same model and solver (31 bodies, 502 contact pairs with self-collision, $4\times 5$\,ms substeps, 10 Newton iterations); other rows use each simulator's own model and solver and are given for reference. \engine's backward megakernel adds $0.56$\,ms on RTX PRO 6000, less than half a forward step and $25.5\times$ faster than MuJoCo Warp's $14.27$\,ms backward. The fused forward+backward step is $6.0$--$11.8\times$ faster across six GPUs ($1.78$ vs.\ $21.0$\,ms on RTX PRO 6000; $8.2$--$11.8\times$ on the four GPUs of \cref{fig:engine}).

\subsection{Accuracy Verification}
\label{supp:engine-accuracy}

\begin{table}[t]
  \centering
  \scriptsize
  \caption{\textbf{Forward accuracy over 16{,}384 held-out Unitree G1 states and VJP accuracy on the 256-state smooth-contact group (G1, 3 cotangents each), against float64 MuJoCo C 3.17 (2000 Newton iterations, tolerance $10^{-13}$).} Backward errors are median relative VJP errors against float64 central finite differences (step $10^{-6}$, floored at the float64 FD precision $\sim 10^{-7}$).}
  \label{tab:supp-acc-overall}
  \setlength{\tabcolsep}{3.5pt}
\resizebox{\linewidth}{!}{%
  \begin{tabular}{@{}lccc@{}}
    \toprule
    & \multicolumn{3}{c}{Gradient (median relative VJP error)} \\
    \cmidrule(lr){2-4}
    Engine (10 Newton iterations, FP32) & Position ($\bar q$) & Velocity ($\bar v$) & Action ($\bar u$) \\
    \midrule
    \textbf{\engine (ours)} & $\mathbf{3.6\times10^{-5}}$ & $\mathbf{1.0\times10^{-5}}$ & $\mathbf{1.9\times10^{-6}}$ \\
    MJX 3.17 (forward-mode Jacobian) & $9.6\times10^{-5}$ & $2.1\times10^{-5}$ & $2.9\times10^{-6}$ \\
    MuJoCo C float64 forward FD (step $10^{-6}$) & $6.7\times10^{-4}$ & $1.5\times10^{-4}$ & $7.8\times10^{-6}$ \\
    Genesis 1.4.5 (velocity only) & --- & $2.3\times10^{-1}$ & --- \\
    MuJoCo Warp, PR \#1535 as opened~\cite{mjwarp_pr1535} & $0.29$ & $0.34$ & $0.015$ \\
    MuJoCo Warp, with follow-up fix~\cite{mjwarp_adjointfix2026} & $5.3\times10^{-5}$ & $1.0\times10^{-5}$ & $2.1\times10^{-6}$ \\
    \midrule
    & \multicolumn{3}{c}{Forward (median relative $v'$ error vs.\ float64 MuJoCo C)} \\
    \midrule
    \textbf{\engine (ours)} & \multicolumn{3}{c}{$\mathbf{8.5\times10^{-7}}$ (0 non-finite of 16{,}384)} \\
    MuJoCo Warp 3.17.0 & \multicolumn{3}{c}{$1.3\times10^{-6}$} \\
    MJX 3.17.0 & \multicolumn{3}{c}{$2.9\times10^{-6}$} \\
    Genesis 1.4.5 & \multicolumn{3}{c}{$5.5\times10^{-2}$} \\
    \bottomrule
  \end{tabular}}%
\end{table}

\begin{table*}[t]
\centering
\scriptsize
\caption{\textbf{Stratified VJP gradient accuracy per contact/actuator group (10 Newton iterations, 256 states $\times$ 3 random cotangent directions per group): median floored relative VJP error for position / velocity / action.} \texttt{n/c}: on G2 (clamped actuators), not comparable: training mode keeps the $-K_v$ damping derivative that exact mode sets to zero.}
\label{tab:supp-acc-grad-groups}
\setlength{\tabcolsep}{5pt}
\resizebox{\textwidth}{!}{%
\begin{tabular}{@{}lllll@{}}
\toprule
Simulator / Mode & G1 (Smooth) & G2 (Clamped actuator) & G3 (Active joint limit) & G4 (Self-contact) \\
\midrule
\textbf{\engine, exact mode} & $\mathbf{3.6\text{e-}5\ /\ 1.0\text{e-}5\ /\ 1.9\text{e-}6}$ & $\mathbf{4.3\text{e-}5\ /\ 1.3\text{e-}5\ /\ 2.0\text{e-}6}$ & $\mathbf{3.8\text{e-}5\ /\ 1.1\text{e-}5\ /\ 1.8\text{e-}6}$ & $\mathbf{3.1\text{e-}5\ /\ 1.0\text{e-}5\ /\ 2.3\text{e-}6}$ \\
\textbf{\engine, training mode} & $\mathbf{3.6\text{e-}5\ /\ 1.0\text{e-}5\ /\ 1.9\text{e-}6}$ & n/c ($8.9\text{e-}3\ /\ 3.4\text{e-}2\ /\ 7.7\text{e-}3$) & $\mathbf{3.8\text{e-}5\ /\ 1.1\text{e-}5\ /\ 1.8\text{e-}6}$ & $\mathbf{3.1\text{e-}5\ /\ 1.0\text{e-}5\ /\ 2.3\text{e-}6}$ \\
MJX 3.17, forward-mode Jacobian & $9.6\text{e-}5\ /\ 2.1\text{e-}5\ /\ 2.9\text{e-}6$ & $1.2\text{e-}4\ /\ 2.3\text{e-}5\ /\ 3.0\text{e-}6$ & $9.4\text{e-}5\ /\ 2.0\text{e-}5\ /\ 3.1\text{e-}6$ & $1.0\text{e-}4\ /\ 2.5\text{e-}5\ /\ 4.8\text{e-}6$ \\
MuJoCo C float64 forward FD & $6.7\text{e-}4\ /\ 1.5\text{e-}4\ /\ 7.8\text{e-}6$ & $7.1\text{e-}4\ /\ 1.8\text{e-}4\ /\ 7.2\text{e-}6$ & $6.6\text{e-}4\ /\ 1.5\text{e-}4\ /\ 7.9\text{e-}6$ & $5.5\text{e-}4\ /\ 1.3\text{e-}4\ /\ 7.8\text{e-}6$ \\
Genesis 1.4.5 (velocity only) & --- / $2.3\text{e-}1$ / --- & --- / $3.4\text{e-}1$ / --- & --- / $2.7\text{e-}1$ / --- & --- / $3.4\text{e-}1$ / --- \\
MuJoCo Warp, PR \#1535 as opened & $2.9\text{e-}1\ /\ 3.4\text{e-}1\ /\ 1.5\text{e-}2$ & $5.0\text{e-}1\ /\ 4.8\text{e-}1\ /\ 8.9\text{e-}2$ & $3.1\text{e-}1\ /\ 3.5\text{e-}1\ /\ 1.8\text{e-}2$ & $3.5\text{e-}1\ /\ 4.0\text{e-}1\ /\ 1.9\text{e-}2$ \\
MuJoCo Warp, follow-up fix~\cite{mjwarp_adjointfix2026} & $5.3\text{e-}5\ /\ 1.0\text{e-}5\ /\ 2.1\text{e-}6$ & n/c & $5.1\text{e-}5\ /\ 1.1\text{e-}5\ /\ 1.9\text{e-}6$ & $4.5\text{e-}5\ /\ 1.0\text{e-}5\ /\ 2.3\text{e-}6$ \\
\bottomrule
\end{tabular}}%
\end{table*}

\paragraph{Verification protocol and training-mode adjoint.}
We evaluate 4 held-out corpora of 4{,}096 G1 states ($16{,}384$ total) against float64 MuJoCo C 3.17 (2000 Newton iterations, tolerance $10^{-13}$) and float64 central finite differences (step $h_{\text{FD}}=10^{-6}$). MuJoCo C groups the states into smooth ground contact (G1, $13{,}480$ states), force-clamped actuators (G2, $476$), active joint limits (G3, $5{,}645$), and robot self-contact (G4, $5{,}668$); a state can belong to several groups. VJPs are checked on 256 states per group with 3 random cotangents each (\cref{tab:supp-acc-overall,tab:supp-acc-grad-groups}). \engine produces zero non-finite outputs or gradients on all 16{,}384 states, and repeated calls from identical inputs give bitwise identical forward states and VJPs on a given GPU and build.

Training mode keeps the $-K_v$ damping derivative on force-clamped actuators (\cref{tab:supp-acc-grad-groups}) and otherwise equals the exact adjoint except for two geometric rules, neither of which changes the forward pass. (1)~For sphere and capsule pairs whose core distance (between centers or closest axis points) is below $1$\,mm, the rotation term $(I - nn^\top)/\nu$ of the contact normal is dropped from the backward pass, keeping the distance derivative. (2)~For capsule--capsule pairs within $2^\circ$ of parallel, both segment parameters are held fixed in the backward pass, so the contact moves with fixed material points. Over $\sim$65k sampled training rows at two checkpoints, the full-batch directional cosine between training-mode and exact adjoints has a median of $1.0000$ (Spearman $\rho = 0.9999$).

\section{Motion Tracking (\method)}
\label{supp:tracking}

\subsection{Task Specifications and Hyperparameters}
\label{supp:tracking-setup}

\begin{table}[t]
\centering
\scriptsize
\caption{\textbf{LAFAN1 tracking reward terms (following BeyondMimic~\cite{liao2026beyondmimic} and SuperMimic~\cite{liao2026supermimic} Table~A.2).} Each exponential tracking kernel is $w \exp(-e / \sigma^2)\,\Delta t$ with $\Delta t = 0.02$\,s, evaluated over 14 tracked bodies (re-anchored horizontally and in yaw to the torso) and the torso anchor. $^{*}$The final recipe trains with the reference velocity taken at each body's center of mass; all reported returns score the body-origin form (reference and robot velocity at the body origin).}
\label{tab:supp-reward}
\setlength{\tabcolsep}{6pt}
\resizebox{\linewidth}{!}{%
\begin{tabular}{@{}lcc@{}}
\toprule
Reward / penalty term & Kernel width $\sigma$ & Weight $w$ \\
\midrule
Body position (14 bodies, anchor-relative) & $0.30$\,m & $+1.0$ \\
Body orientation (14 bodies, anchor-relative) & $0.40$\,rad & $+1.0$ \\
Body linear velocity$^{*}$ & $1.00$\,m/s & $+1.0$ \\
Body angular velocity & $3.14$\,rad/s & $+1.0$ \\
Anchor position (torso) & $0.30$\,m & $+0.5$ \\
Anchor orientation (torso) & $0.40$\,rad & $+0.5$ \\
Action rate penalty $\lVert u_t - u_{t-1}\rVert^2$ & --- & $-0.1$ \\
Joint limit violation penalty & --- & $-2.0$ \\
\bottomrule
\end{tabular}}%
\end{table}

\begin{table}[t]
\centering
\scriptsize
\caption{\textbf{\method hyperparameters for 40-clip LAFAN1 tracking on the 29-DoF Unitree G1.}}
\label{tab:supp-hyper}
\setlength{\tabcolsep}{4pt}
\begin{tabular}{@{}lp{0.64\linewidth}@{}}
\toprule
Hyperparameter & Value \\
\midrule
Parallel windows & $B = 2048$ BPTT training windows per update ($4096$ in the earlier recipe) \\
Window horizon & $H = 64$ steps ($1.28$\,s at $50$\,Hz; $4\times 5$\,ms physics substeps/step) \\
Discount & $\gamma = 0.99$, $\mathrm{TD}(\lambda_{\text{TD}} = 0.95)$ \\
Actor network & Goal encoder--decoder MLP $2048$--$1024$--$512$--$256$ ($\to 32$-D $\tanh$ latent), LayerNorm + SiLU, \textmu P initialization; outputs joint offsets around the reference frame \\
Critic network & Privileged MLP $256$--$128$, LayerNorm + SiLU, fit on pre-transition states \\
Optimizers & Actor hidden matrices: Muon (RMS step $0.1$); all other parameters: Adam \\
Learning rate & $2\times 10^{-4}$; 100-update warmup; actor $3\times$ for updates 0--500, ramping back to $1\times$ by update 800; constant until update 1{,}000, then cosine decay to $0.1\times$ over 16{,}000 updates; policy LR capped at $0.02\times$ from update 16{,}000 \\
Action targets & $\delta = 0.03$\,rad per-joint RMS (decays to $0.3\times\delta$ with LR); $M = 8$ regression minibatch steps per update, each on $16{,}384$ targets sampled with replacement from the $N$ window samples (together about one pass over the $N$ samples in expectation); separate Adam, lr $6\times 10^{-4}$ \\
Per-sample clip & $\kappa = 3\times$ batch median of $\lVert g_i \rVert_2$ (applied at each backward step and to the window output; window output only performs the same, \cref{tab:supp-abl-final}) \\
Actor EMA & $\beta_n = \min(0.999, (1+n)/(10+n))$ \\
Exploration noise & $\sigma = 0.1$ in window rollouts ($0.05$ in the earlier recipe); $\sigma_{\text{col}} = 0.1$ in collector \\
Replay collector & $4096$ envs, $\text{UTD} = 0.4$, FIFO buffer of $1000\times 4096$ states; return-weighted reset bins + $15\%$ pre-fall bin sampling every 150 updates \\
Domain rand. & Per episode, training only: foot friction $\mathcal{U}[0.3, 1.2]$; joint encoder bias $\mathcal{U}[-0.01, 0.01]$\,rad; torso CoM offset $\mathcal{U}$ within $\pm(0.025, 0.05, 0.05)$\,m; uniform observation noise with half-widths $0.01$\,rad (joint position), $0.5$\,rad/s (joint velocity), $0.2$\,rad/s (base angular velocity), $0.05$ (projected gravity), $0.05$ (each entry of the 6-D anchor orientation); reset noise on root pose ($\pm(0.05, 0.05, 0.01)$\,m, $\pm(0.1, 0.1, 0.2)$\,rad), root velocity, and joints ($\pm 0.1$\,rad) \\
Update time & median $0.240$\,s/update on one RTX PRO 6000; 24{,}000 updates in $1.66$\,h \\
\bottomrule
\end{tabular}
\end{table}

\Cref{tab:supp-reward,tab:supp-hyper} list the complete reward terms, observations, domain randomization ranges, and hyperparameters for 40-clip LAFAN1 tracking. The actor observes a 10-frame proprioception history ($10\times 93$-D: 29 joint positions relative to default, 29 joint velocities, base angular velocity, projected gravity, and the previous action as applied, i.e.\ clipped to the setpoint range; uniform noise during training as in \cref{tab:supp-hyper}), a 10-frame future reference window ($10\times 68$-D in the reference's own frame), and the 6-D relative orientation between the reference anchor and robot anchor (with uniform noise of half-width $0.05$ during training; no global root position is observed by the actor, enabling blind long-horizon evaluation). A fall is a termination under the pelvis-height and tilt thresholds of \cref{supp:comp-graph} ($>0.25$\,m, $>0.8$\,rad).

\paragraph{Action space.} The tracking actor outputs joint offsets around the reference motion: the PD setpoint is the next reference frame's joint angles plus the scaled action, $u_t = q^{\text{ref}}_{t+1} + \alpha\, a_t$, as in SuperTrack~\cite{fussell2021supertrack} and InstantTrack~\cite{dao2026instanttrack}. BeyondMimic and SuperMimic use offsets around the default pose, $u_t = q^0 + \alpha\, a_t$. On an earlier recipe, switching from default-pose offsets to reference offsets raised the return by $0.008$ (\cref{tab:supp-abl-abs}). Tasks without a reference motion (velocity tracking, terrain traversal) use default-pose offsets, and the DeepMind Control and Brax tasks use their native action spaces.

\paragraph{Baselines.} SuperMimic is run from its official release (commit \texttt{abc124c}) with proprioceptive actor input, 10-frame observation history, and all 40 LAFAN1 clips, seeds 0--2, a $\sim 3.2$\,h budget, and other settings as released; InstantTrack v1.4.0 (commit \texttt{b9512a9}) is run with its released settings, with \engine as its simulator on the shared G1 model. Both are trained and evaluated with all G1 collision pairs, including self-collisions, enabled.

\subsection{Full LAFAN1 Results}
\label{supp:lafan-results}

\begin{table*}[t]
\centering
\scriptsize
\caption{\textbf{Per-seed plateau returns, per-seed threshold-crossing times (mean $\pm$ s.d.\ across individual seeds), and cold-start vs.\ warm-start compilation overheads on 40-clip LAFAN1 tracking (MuJoCo Warp).} The first time in each cell is the first crossing of the seed-mean curve, as in \cref{tab:levels}; parentheses give the per-seed mean $\pm$ s.d.\ and the number of seeds that reach the level. Times measured on other GPUs are converted to the RTX PRO 6000 by update count.}
\label{tab:supp-plateaus}
\setlength{\tabcolsep}{3.8pt}
\resizebox{\textwidth}{!}{%
\begin{tabular}{@{}lcccccc@{}}
\toprule
Method (RTX PRO 6000) & Individual seed plateaus (last 30\,min) & Time to 0.721 (min) & Time to 0.866 (min) & Time to 0.890 (min) & Time to 0.929 (min) & Startup / compile overhead \\
\midrule
\method (5 seeds) & $0.9291, 0.9308, 0.9309, 0.9290, 0.9266$ & $1.5$ ($1.5\pm 0.1$, 5/5) & $5.3$ ($5.3\pm 0.4$, 5/5) & $17.2$ ($16.1\pm 2.1$, 5/5) & $78.1$ ($80.2\pm 7.7$, 4/5) & actor compile $37$\,s cold, $7$\,s cached \\
\method, earlier recipe (3 seeds) & $0.9089, 0.9153, 0.9171$ ($0.914\pm 0.004$) & $3.1$ ($3.1\pm 0.1$, 3/3) & $19.2$ ($19.6\pm 2.8$, 3/3) & $55.1$ ($65.6\pm 36.5$, 3/3) & Not reached & --- \\
InstantTrack v1.4 (3 seeds, on \engine) & $0.8908, 0.8903, 0.8895$ ($0.890\pm 0.001$) & $1.5$ ($1.4\pm 0.1$, 3/3) & $33.2$ ($31.7\pm 1.5$, 3/3) & $155.3$ ($156.2\pm 2.0$, 3/3) & Not reached & $107$--$156$\,s warm ($539$\,s cold) \\
SuperMimic (official code, 3 seeds) & $0.851, 0.850, 0.815$ ($0.838\pm 0.021$) & $16.1$ ($15.7\pm 2.7$, 3/3) & Not reached (max $0.852$) & Not reached & Not reached & $70$\,s warm ($496$\,s cold) \\
PPO, BeyondMimic config (1 seed, 6\,h) & $0.721$ (best checkpoint $0.744$; final $0.697$) & $298.4$ (1/1) & Not reached & Not reached & Not reached & $\sim 3$\,s first-step compile \\
PPO, BeyondMimic config (3 seeds, 48\,h)\textsuperscript{a} & $0.7766, 0.7873, 0.7815$ ($0.782\pm 0.005$) & $266$ ($240\pm 19$, 3/3) & Not reached & Not reached & Not reached & $\sim 3$\,s first-step compile \\
\bottomrule
\end{tabular}}%
\par\smallskip\raggedright\scriptsize
\textsuperscript{a}\,48\,h of training on RTX 3090 (one seed finished on an A40); times converted to the RTX PRO 6000 by update count (0.1227\,s/update), so the runs span 13.6--16.8\,h.
\end{table*}

\begin{table*}[t]
\centering
\scriptsize
\caption{\textbf{LAFAN1 results for all runs (mean $\pm$ s.d.\ over seeds).} Final return is the last evaluation of each run, or the last-30-min plateau where marked (SHAC's last-30-min mean: $0.817$). All returns are scored in deterministic MuJoCo Warp. With \method, all 5 seeds first reach $0/40$ failed clips at update 1{,}250 ($5.3$\,min) and have $0/40$ at every checkpoint from update 13{,}750 ($57$\,min) on (from $32$\,min when scored in \engine).}
\label{tab:supp-lafan-full}
\setlength{\tabcolsep}{3.5pt}
\resizebox{\textwidth}{!}{%
\begin{tabular}{@{}lllcccccccc@{}}
\toprule
Method & Operator & GPU & Seeds & Return @0.5\,h & Return @1.0\,h & Return @2.0\,h & Return @3.0\,h & Time to 0.82 (h) & Final return (budget) & Failed clips (/40) \\
\midrule
\method (final recipe) & \engine & RTX PRO 6000 & 5 & $\mathbf{0.908\pm 0.002}$ & $\mathbf{0.923\pm 0.002}$ & --- & --- & \textbf{0.04\,h} (2.5\,min) & $\mathbf{0.930\pm 0.004}$ (1.66\,h) & $\mathbf{0.0\pm 0.0}$ \\
\method (earlier recipe) & \engine & RTX PRO 6000 & 3 & $0.872\pm 0.006$ & $0.892\pm 0.011$ & $0.905\pm 0.016$ & $0.916\pm 0.003$ & 0.12\,h & $0.917\pm 0.002$ (3.07\,h) & $0.0\pm 0.0$ \\
\method (absolute actions) & \engine & RTX PRO 6000 & 3 & $0.829\pm 0.005$ & $0.850\pm 0.004$ & $0.867\pm 0.001$ & $0.870\pm 0.002$ & 0.48\,h & $0.871\pm 0.003$ (3.14\,h) & $0.0\pm 0.0$ \\
\method (absolute actions) & \engine & H100 & 1 & $0.820$ & $0.854$ & $0.867$ & $0.868$ & 0.40\,h & $0.864$ (3.10\,h) & $0.0$ \\
\method (absolute actions) & MuJoCo Warp & RTX PRO 6000 & 3 & $0.662\pm 0.016$ & $0.743\pm 0.040$ & $0.803\pm 0.007$ & $0.831\pm 0.009$ & 2.39\,h & $0.830\pm 0.011$ (3.00\,h) & $0.7\pm 0.6$ \\
InstantTrack v1.4 & \engine & RTX PRO 6000 & 3 & $0.863\pm 0.001$ & $0.872\pm 0.001$ & $0.885\pm 0.001$ & $0.892\pm 0.003$ & \textbf{0.04\,h} & $0.890\pm 0.001$ (plateau) & $0.0\pm 0.0$ \\
SuperMimic (official code) & MuJoCo Warp & RTX / H100 & 3 & --- & --- & --- & $0.847$ & 0.76\,h & $0.838\pm 0.021$ (plateau) & $0.9\pm 1.5$ \\
SHAC (guard relaxed) & \engine & RTX PRO 6000 & 3 & $0.727\pm 0.017$ & $0.792\pm 0.010$ & $0.810\pm 0.002$ & $0.820\pm 0.006$ & 2.60\,h & $0.820\pm 0.010$ (3.13\,h) & $1.7\pm 1.5$ \\
SHAC (guard relaxed) & MuJoCo Warp & RTX PRO 6000 & 1 & $0.493$ & $0.602$ & $0.707$ & $0.757$ & Not reached & $0.758$ (3.13\,h) & $5.0$ \\
SHAC (published guard) & \engine & RTX PRO 6000 & 3 & \multicolumn{4}{c}{Halts at $2.5$--$5.2$\,min on $10^6$ gradient guard} & Not reached & $0.384\pm 0.102$ (0.06\,h) & $22.7\pm 4.2$ \\
PPO (BeyondMimic config) & \engine (fwd) & RTX PRO 6000 & 1 & $0.454$ & $0.544$ & $0.623$ & $0.635$ & Not reached & $0.697$ final (6.00\,h); plateau $0.721$, first reached at 298\,min; best checkpoint $0.744$ (345\,min) & $8.0$ \\
PPO (SuperMimic actor/resets) & \engine (fwd) & RTX PRO 6000 & 1 & $0.428$ & $0.477$ & $0.580$ & $0.643$ & Not reached & $0.662$ (6.00\,h) & $5.0$ \\
\bottomrule
\end{tabular}}%
\end{table*}

Training times exclude one-time compilation, listed per method in \cref{tab:supp-plateaus}.

\subsection{Cross-Simulator Re-Scoring and SuperMimic Runs}
\label{supp:baseline-runs}

Identical checkpoints score within $0.014$ in the two simulators, and identically once training has converged (\cref{tab:supp-sim2sim}). \Cref{tab:supp-supermimic} lists per-seed training speed and return of the official SuperMimic runs.

\begin{table}[t]
\centering
\scriptsize
\caption{\textbf{Cross-simulator transfer (\texttt{sim2sim}): identical checkpoints evaluated in deterministic MuJoCo Warp and in \engine (mean $\pm$ s.d.\ over 3 seeds, 5 for the final recipe, 10 Newton iterations, self-collision enabled).} }
\label{tab:supp-sim2sim}
\setlength{\tabcolsep}{3.5pt}
\resizebox{\linewidth}{!}{%
\begin{tabular}{@{}llcccc@{}}
\toprule
Policy recipe & Trained in & Updates & MuJoCo Warp & \engine & $\Delta$ (NE $-$ MJW) \\
\midrule
\method (absolute) & \engine & 6{,}000 & $0.809\pm 0.017$ & $0.819\pm 0.018$ & $+0.010$ \\
\method (absolute) & MuJoCo Warp & 6{,}000 & $0.830\pm 0.011$ & $0.817\pm 0.005$ & $-0.014$ \\
\method (absolute) & \engine & 29{,}000 & $0.872\pm 0.005$ & $0.872\pm 0.005$ & $0.000$ \\
\method (earlier recipe) & \engine & 6{,}000 & $0.878\pm 0.012$ & $0.878\pm 0.012$ & $0.000$ \\
\method (earlier recipe) & \engine & 29{,}000 & $0.917\pm 0.002$ & $0.917\pm 0.002$ & $0.000$ \\
\method (final recipe) & \engine & 24{,}000 & $0.930\pm 0.004$ & $0.930\pm 0.003$ & $0.000$ \\
\bottomrule
\end{tabular}}%
\end{table}

\begin{table}[t]
\centering
\scriptsize
\caption{\textbf{Official SuperMimic (commit \texttt{abc124c}), 3 seeds: per-seed training speed and return.} Times measured on H100 are converted to the RTX PRO 6000 by update count ($0.315$\,s/update). Per-seed times are the first evaluation at or above the level; the last row is the first crossing of the seed-mean curve (as in \cref{tab:levels}). Plateau: last-30-min mean, scored in MuJoCo Warp.}
\label{tab:supp-supermimic}
\setlength{\tabcolsep}{3pt}
\resizebox{\linewidth}{!}{%
\begin{tabular}{@{}lccccc@{}}
\toprule
Seed (GPU) & Steady s/update & To 0.721 (min) & To 0.82 (min) & Plateau & Peak \\
\midrule
Seed 0 (H100) & $0.285$ & $18.4$ & $47.2$ & $0.851$ & $0.852$ \\
Seed 1 (RTX PRO 6000) & $0.317$ & $13.1$ & $42.0$ & $0.850$ & $0.851$ \\
Seed 2 (H100) & $0.281$ & $15.7$ & $42.0$ & $0.815$\textsuperscript{a} & $0.848$ \\
\midrule
Mean $\pm$ s.d. & --- & $15.7\pm 2.7$ & $43.7\pm 3.0$ & $0.838\pm 0.021$ & $0.850\pm 0.002$ \\
Seed-mean curve & --- & $16.1$ & $45.5$ & --- & --- \\
\bottomrule
\end{tabular}}%
\par\smallskip\raggedright\scriptsize
\textsuperscript{a}\,Seed 2 peaked at $0.847$ (168\,min); its final checkpoint fails $2.7$ clips on average (three evaluations: 2, 4, 2).
\end{table}

\subsection{Tracking Metrics and Physical Deployment}
\label{supp:mpjpe-real}

\begin{table}[t]
\centering
\scriptsize
\caption{\textbf{Tracking metrics on 40-clip LAFAN1 at each run's final checkpoint (MuJoCo Warp, deterministic, each clip played once from frame 0).} Return: full-clip normalized return. Success: fraction of clip rollouts completed without a fall. MPJPE: mean 14-body anchor-relative position error. Mean $\pm$ s.d.\ over seeds.}
\label{tab:supp-robustness}
\setlength{\tabcolsep}{3pt}
\begin{tabular}{@{}lcccc@{}}
\toprule
Method & Seeds & Return $\uparrow$ & Success $\uparrow$ & MPJPE (cm) $\downarrow$ \\
\midrule
\textbf{\method (ours)} & 5 & $\mathbf{0.930}\pm 0.004$ & $\mathbf{100.0\%}$ & $\mathbf{2.16}\pm 0.02$ \\
InstantTrack v1.4 & 3 & $0.893\pm 0.002$ & $\mathbf{100.0\%}$ & $2.57\pm 0.05$ \\
SuperMimic (official code) & 3 & $0.819\pm 0.057$ & $97.5\%$ & $3.45\pm 0.83$ \\
\bottomrule
\end{tabular}
\end{table}

\begin{table}[t]
\centering
\scriptsize
\caption{\textbf{Long-horizon blind root drift over the 30 LAFAN1 clips longer than $200$\,s ($225.7$--$273.1$\,s; evaluated in MuJoCo Warp without root-position observation).}}
\label{tab:supp-drift}
\setlength{\tabcolsep}{4pt}
\resizebox{\linewidth}{!}{%
\begin{tabular}{@{}lcccc@{}}
\toprule
Method & Seeds & Mean root drift (m) & End-of-clip drift (m) & Falls (clips $\times$ seeds) \\
\midrule
\textbf{\method (ours)} & 5 & $\mathbf{0.154\pm 0.018}$ & $\mathbf{0.270\pm 0.034}$ & \textbf{0 / 150 (0.0\%)} \\
InstantTrack v1.4 & 3 & $0.316\pm 0.022$ ($2.1\times$) & $0.537\pm 0.063$ ($2.0\times$) & 0 / 90 (0.0\%) \\
SuperMimic (official code) & 3 & $1.167\pm 0.564$ ($7.6\times$) & $1.871\pm 0.827$ ($6.9\times$) & 2 / 90 (2.2\%) \\
\bottomrule
\end{tabular}}%
\end{table}

\begin{table*}[t]
\centering
\scriptsize
\caption{\textbf{Physical Unitree G1 deployment: onboard 50\,Hz tracking telemetry over \realTrialCount{} consecutive trials of \realClipCount{} LAFAN1 clips (\realTotalFrames{} frames, \realTotalSec\,s; 3.0\,h earlier-recipe EMA checkpoint, simulation return $0.911$, no real-world fine-tuning).} Joint RMS error measures $\sqrt{\tfrac{1}{Td}\sum \|q - q^{\text{ref}}\|_2^2}$ across all 29 actuated joints; torso tilt measures IMU projected-gravity tilt error during clip playback; onboard kernels report logged 50\,Hz body position / body orientation / root orientation tracking scores.}
\label{tab:supp-real}
\setlength{\tabcolsep}{4pt}
\resizebox{\textwidth}{!}{%
\begin{tabular}{@{}llcccccc@{}}
\toprule
LAFAN1 clip & Motion type & Trial (dur.) & Joint RMS ($^\circ$) $\downarrow$ & Torso tilt mean / p95 ($^\circ$) $\downarrow$ & Onboard kernels (pos / ori / root) $\uparrow$ & $\tau/\tau_{\max}$ p99 & Falls \\
\midrule
\texttt{walk1\_subject2} & Locomotion / turns & Trial 1 (18.5\,s) & $3.78^\circ$ & $2.66^\circ\ /\ 4.8^\circ$ & $0.995\ /\ 0.949\ /\ 0.982$ & $0.36$ & 0 \\
\texttt{walk1\_subject2} & Locomotion / turns & Trial 2 (18.5\,s) & $3.70^\circ$ & $2.51^\circ\ /\ 4.8^\circ$ & $0.996\ /\ 0.951\ /\ 0.985$ & $0.37$ & 0 \\
\texttt{dance1\_subject3} & Agile whole-body dance & Trial 1 (15.4\,s) & $4.62^\circ$ & $3.56^\circ\ /\ 5.9^\circ$ & $0.994\ /\ 0.939\ /\ 0.989$ & $0.35$ & 0 \\
\texttt{dance2\_subject5} & Rhythmic turns & Trial 1 (30.0\,s) & $3.19^\circ$ & $2.62^\circ\ /\ 4.8^\circ$ & $0.994\ /\ 0.963\ /\ 0.988$ & $0.47$ & 0 \\
\texttt{dance2\_subject3} & Fast spins / wide steps & Trial 1 (35.3\,s) & $3.95^\circ$ & $3.09^\circ\ /\ 6.9^\circ$ & $0.992\ /\ 0.945\ /\ 0.967$ & $0.32$ & 0 \\
\texttt{fightAndSports1\_subject4} & Dynamic punches / kicks & Trial 1 (8.8\,s) & $4.23^\circ$ & $2.00^\circ\ /\ 3.7^\circ$ & $0.994\ /\ 0.945\ /\ 0.991$ & $0.37$ & 0 \\
\texttt{fightAndSports1\_subject4} & Dynamic punches / kicks & Trial 2 (8.8\,s) & $4.44^\circ$ & $2.15^\circ\ /\ 4.0^\circ$ & $0.992\ /\ 0.942\ /\ 0.991$ & $0.38$ & 0 \\
\midrule
\textbf{All 5 clips (7 trials)} & \textbf{6,762 frames @ 50\,Hz} & \textbf{135.3\,s} & $\mathbf{3.88^\circ}$ & $\mathbf{2.77^\circ}$ (3.7--6.9$^\circ$) & $\mathbf{0.994\ /\ 0.949\ /\ 0.982}$ & $\le 0.47$ & \textbf{0 / 7} \\
\bottomrule
\end{tabular}}%
\par\smallskip\raggedright\scriptsize
Onboard 50\,Hz controller telemetry across all \realTrialCount{} runs: control period median \realCtrlDtMed\,ms (max \realCtrlDtMax\,ms), state estimation age median \realStateAgeMed\,ms (99th percentile $\le \realStateAgePninetyfivePlus$\,ms), and actuator setpoint saturation fraction \realSatCtrlMax.
\end{table*}

\paragraph{MPJPE, success rate, and long-horizon drift (\cref{tab:supp-robustness,tab:supp-drift}).}
On fall-free clip success and 14-body MPJPE (\cref{tab:supp-robustness}), \method completes all 40 clips on every seed with $2.16$\,cm MPJPE, against $2.57$\,cm for InstantTrack and $3.45$\,cm for SuperMimic's official code. Over the 30 clips exceeding $200$\,s evaluated blind (\cref{tab:supp-drift}), \method has $2.0$--$2.1\times$ lower mean and end-of-clip root drift than InstantTrack and $6.9$--$7.6\times$ lower than SuperMimic.

\paragraph{Physical Unitree G1 deployment (\cref{tab:supp-real}).}
Two checkpoints from earlier recipes than the final one in \cref{tab:levels} ran the same seven trials over five LAFAN1 clips without a fall: a policy trained for 9.8\,min on one RTX PRO 6000 (simulation return $0.867$) and a policy trained for 3.0\,h (simulation return $0.911$, $0/40$ falls). The telemetry in \cref{tab:supp-real} was logged with the 3.0\,h policy.

Both run as EMA actors at 50\,Hz onboard a 29-DoF Unitree G1, with no real-world data, system identification, fine-tuning or reward retuning; transfer relies on training-time domain randomization (\cref{tab:supp-hyper}).
Each of the \realTrialCount{} trials (\realClipCount{} clips, \realTotalFrames{} active frames, \cref{tab:supp-real}) started the robot in a standing pose, transitioned into the reference clip, and ran under a slack overhead safety strap, with 0/\realTrialCount{} falls. Across all \realTotalFrames{} active 50\,Hz frames (\realTotalSec\,s total), the frame-weighted 29-joint position RMS error is $\realJointRmsWeighted^\circ$ (range $\realJointRmsMin^\circ$--$\realJointRmsMax^\circ$ across trials), the frame-weighted mean torso tilt error is $\realTiltWeighted^\circ$ (95th percentile $\realTiltPninetyfiveMin^\circ$--$\realTiltPninetyfiveMax^\circ$, overall clip max $\realTiltMaxClip^\circ$), and the onboard tracking kernels are \realKPosWeighted{} (body pos), \realKOriWeighted{} (body ori), \realKAngWeighted{} (ang.\ vel.), and \realKRootWeighted{} (root ori), with 99th-percentile joint torque utilization $\le \realTauPninetyfiveMax$ and $\realSatCtrlMax$ actuator setpoint saturation.

\FloatBarrier
\section{DeepMind Control and Brax Benchmarks}
\label{supp:dmc}

\subsection{Per-Task \napo vs.\ Fixed \napo$^\dagger$ Configurations and Full Selection Grid}
\label{supp:dmc-configs}

\begin{table*}[t]
\centering
\scriptsize
\caption{\textbf{Complete per-task configuration table for \napo vs.\ fixed-configuration \napo$^\dagger$ across all 16 DeepMind Control and Brax benchmarks ($10^8$ environment steps, evaluated over 10 deterministic episodes in canonical \texttt{dm\_control} 1.0.51 or Brax MJX).} Fixed \napo$^\dagger$ uses one setting on all 16 tasks: $H=8$, $B=512$ environments, pathwise estimator (\texttt{pw}), $\sigma=0.2$, Adam ($\text{lr}=6\times 10^{-4}$, warmup 0), straight-through action clamp (\texttt{ste}), and EMA $0.999$ (5 seeds; 3 on Brax ant, hopper, and walker2d). Per-task \napo selects from the 8-arm estimator $\times\ \sigma$ grid in \cref{tab:supp-dmc-grid} (and $H\in\{16,32\}, B=1024$ on the 3 tasks with long-horizon contacts or sparse rewards). Final return is the last checkpoint of the budget.}
\label{tab:supp-dmc-configs}
\setlength{\tabcolsep}{4pt}
\resizebox{\textwidth}{!}{%
\begin{tabular}{@{}lccccccl@{}}
\toprule
Task & $H$ & $B$ & $\sigma$ & Grad.\ est.\ & Per-task \napo ($n$) & Fixed \napo$^\dagger$ ($n=3$--$5$) & Note / $4\times 10^8$-step extended run \\
\midrule
\texttt{acrobot-swingup} & 8 & 512 & 0.05 & \texttt{pwall} & $\mathbf{250\pm 21}$ (3) & $237\pm 48$ & $4\times 10^8$ steps ($H=128$): $\mathbf{484\pm 36}$ \\
\texttt{acrobot-swingup-sparse} & 8 & 512 & 0.20 & \texttt{blend} & $\mathbf{104\pm 34}$ (5) & $77\pm 12$ & $4\times 10^8$ steps ($H=128$): $\mathbf{258\pm 136}$ \\
\texttt{cartpole-swingup} & 8 & 512 & 0.10 & \texttt{blend} & $\mathbf{883\pm 0}$ (3) & $\mathbf{883\pm 0}$ & All 8 grid arms score $868$--$883$ \\
\texttt{cartpole-swingup-sparse} & 32 & 1024 & 0.40 & \texttt{pw} & $\mathbf{675\pm 139}$ (3) & $0$ ($292$ w/\ \texttt{blend}) & $H=32$, $\sigma=0.05$: $617\pm 346$ (5 seeds, $4/5$ solved) \\
\texttt{cheetah-run} & 8 & 512 & 0.20 & \texttt{pwall} & $\mathbf{919\pm 4}$ (3) & $917\pm 5$ & --- \\
\texttt{finger-spin} & 8 & 512 & 0.10 & \texttt{pw} & $955\pm 11$ (3) & $945\pm 21$ & --- \\
\texttt{hopper-hop} & 8 & 512 & 0.05 & \texttt{pw} & $\mathbf{395\pm 64}$ (3) & $333\pm 21$ & $4\times 10^8$ steps: $\mathbf{497\pm 83}$ \\
\texttt{hopper-stand} & 8 & 512 & 0.05 & \texttt{pwall} & $\mathbf{939\pm 16}$ (3) & $839\pm 252$ & --- \\
\texttt{pendulum-swingup} & 8 & 512 & 0.10 & \texttt{blend} & $\mathbf{792\pm 4}$ (3) & $781\pm 15$ & All 8 grid arms score $769$--$792$ \\
\texttt{reacher-hard} & 8 & 512 & 0.20 & \texttt{pwall} & $974\pm 4$ (3) & $968\pm 3$ & --- \\
\texttt{walker-run} & 8 & 512 & 0.10 & \texttt{pw} & $\mathbf{904\pm 50}$ (3) & $865\pm 45$ & --- \\
\texttt{walker-walk} & 8 & 512 & 0.20 & \texttt{blend} & $\mathbf{981\pm 1}$ (5) & $980\pm 2$ & All 8 grid arms score $977$--$981$ \\
\midrule
\texttt{Brax ant} & 16 & 1024 & 0.10 & \texttt{blend} & $\mathbf{7019\pm 666}$ (3) & $650\pm 176$ & \texttt{blend}, $\sigma=0.2$ ($H=8$, $B=512$): $5881$ \\
\texttt{Brax halfcheetah} & 8 & 512 & 0.05 & \texttt{pwall} & $\mathbf{17689\pm 235}$ (3) & $15623\pm 994$ & --- \\
\texttt{Brax hopper} & 8 & 512 & 0.20 & \texttt{pw} & $2716\pm 518$ (3) & $2716\pm 518$ & --- \\
\texttt{Brax walker2d} & 16 & 1024 & 0.40 & \texttt{pw} & $\mathbf{6725\pm 203}$ (3) & $5918\pm 656$ & \texttt{blend}, $\sigma=0.2$ ($H=8$, $B=512$): $6418$ \\
\bottomrule
\end{tabular}}%
\end{table*}

\begin{table*}[t]
\centering
\scriptsize
\caption{\textbf{Full 8-arm selection round across the 12 DeepMind Control tasks and Brax HalfCheetah ($H=8, B=512$, Adam, $10^8$ steps, 3--5 seeds per cell).} Estimators: \texttt{pw} (pathwise), \texttt{pwall} (pathwise + regress all window steps), and \texttt{blend} (Suh \etal~\cite{suh2022dodiff} variance-weighted combination of pathwise and antithetic likelihood-ratio gradients).}
\label{tab:supp-dmc-grid}
\setlength{\tabcolsep}{4.5pt}
\resizebox{\textwidth}{!}{%
\begin{tabular}{@{}lcccccccc@{}}
\toprule
Task & \texttt{pw} $\sigma=0.2$ (\napo$^\dagger$) & \texttt{pw} $\sigma=0.1$ & \texttt{pw} $\sigma=0.05$ & \texttt{blend} $\sigma=0.2$ & \texttt{blend} $\sigma=0.1$ & \texttt{blend} $\sigma=0.05$ & \texttt{pwall} $\sigma=0.2$ & \texttt{pwall} $\sigma=0.05$ \\
\midrule
\texttt{acrobot-swingup} & 237 & 195 & 228 & 220 & 213 & 218 & 208 & \textbf{250} \\
\texttt{acrobot-swingup-sparse} & 77 & 85 & 77 & \textbf{104} & 77 & 81 & 90 & 52 \\
\texttt{cartpole-swingup} & \textbf{883} & 882 & 872 & 876 & \textbf{883} & 868 & 878 & 880 \\
\texttt{cartpole-swingup-sparse} & 0 & 0 & 0 & \textbf{292} & 0 & 0 & 0 & 6 \\
\texttt{cheetah-run} & 917 & 908 & 911 & 918 & 916 & 805 & \textbf{919} & 907 \\
\texttt{finger-spin} & 945 & \textbf{955} & 797 & 793 & 737 & 769 & 705 & 850 \\
\texttt{hopper-hop} & 333 & 257 & \textbf{395} & 303 & 365 & 375 & 338 & 191 \\
\texttt{hopper-stand} & 839 & 754 & 914 & 723 & 753 & 666 & 764 & \textbf{939} \\
\texttt{pendulum-swingup} & 781 & 783 & 784 & 792 & \textbf{792} & 786 & 769 & 770 \\
\texttt{reacher-hard} & 968 & 969 & 883 & 966 & 956 & 958 & \textbf{974} & 958 \\
\texttt{walker-run} & 865 & \textbf{904} & 887 & 861 & 865 & 857 & 850 & 884 \\
\texttt{walker-walk} & 980 & 977 & 980 & \textbf{981} & 978 & 978 & 980 & 980 \\
\texttt{Brax halfcheetah} & 15623 & 17379 & 17057 & 15823 & 13840 & 10627 & 16205 & \textbf{17689} \\
\bottomrule
\end{tabular}}%
\end{table*}

\begin{table*}[t]
\centering
\scriptsize
\caption{\textbf{Exclusive single-GPU (V100) wall-clock seconds (including JIT compilation) to reach $0.8\,R^*$ on the 12 DMC tasks and Brax HalfCheetah.} $R^*$ is the best mean final return among single-configuration methods, as in \cref{tab:dmc}. \texttt{n/r}: $0.8\,R^*$ not reached within budget. Last row: tasks on which \napo$^\dagger$ is faster.}
\label{tab:supp-dmc-timed}
\setlength{\tabcolsep}{4.5pt}
\resizebox{\textwidth}{!}{%
\begin{tabular}{@{}lcccccccc@{}}
\toprule
Task & $0.8\,R^*$ target & \napo$^\dagger$ (\texttt{pw}, $\sigma=0.2$) & PPO~NE & SAC~NE & PPO~orig & SAC~orig & SHAC & AHAC \\
\midrule
\texttt{acrobot-swingup} & 189.7 & 298.9\,s & \textbf{215.2\,s} & n/r & n/r & n/r & n/r & n/r \\
\texttt{acrobot-swingup-sparse} & 83.2 & \textbf{301.3\,s} & n/r & n/r & n/r & n/r & n/r & n/r \\
\texttt{cartpole-swingup} & 706.0 & \textbf{45.2\,s} & 126.5\,s & 106.3\,s & 138.0\,s & 79.2\,s & n/r & n/r \\
\texttt{cartpole-swingup-sparse}\textsuperscript{a} & 431.1 & 311.8\,s\textsuperscript{a} & n/r\textsuperscript{a} & \textbf{188.3\,s} & n/r & 317.4\,s & n/r & n/r \\
\texttt{cheetah-run} & 734.0 & \textbf{38.9\,s} & n/r & 280.9\,s & 231.4\,s & 226.4\,s & n/r & n/r \\
\texttt{finger-spin} & 769.2 & 203.3\,s & 1077.1\,s & 744.3\,s & n/r & n/r & n/r & \textbf{61.9\,s}\textsuperscript{b} \\
\texttt{hopper-hop} & 266.2 & \textbf{88.1\,s} & n/r & n/r & n/r & n/r & n/r & n/r \\
\texttt{hopper-stand} & 671.2 & \textbf{77.7\,s} & n/r & 271.9\,s & n/r & 387.1\,s & n/r & n/r \\
\texttt{pendulum-swingup} & 633.5 & \textbf{28.1\,s} & 61.5\,s & 114.1\,s & 70.7\,s & n/r & n/r & n/r \\
\texttt{reacher-hard} & 783.0 & \textbf{34.1\,s} & 82.7\,s & 58.7\,s & 421.9\,s & 86.1\,s & n/r & n/r \\
\texttt{walker-run} & 691.8 & \textbf{98.2\,s} & 495.4\,s & 293.1\,s & n/r & 1275.5\,s & n/r & n/r \\
\texttt{walker-walk} & 784.7 & \textbf{72.3\,s} & 209.8\,s & 143.5\,s & 439.8\,s & 153.4\,s & n/r & n/r \\
\texttt{Brax halfcheetah} & 12658.4 & \textbf{189.0\,s} & n/r & n/r & n/r & n/r & n/r & n/r \\
\midrule
\textbf{\napo$^\dagger$ faster} & --- & --- & \textbf{12 / 13} & \textbf{12 / 13} & \textbf{13 / 13} & \textbf{13 / 13} & \textbf{13 / 13} & \textbf{12 / 13} \\
\bottomrule
\end{tabular}}%
\par\smallskip\raggedright\scriptsize
\textsuperscript{a}\,PPO~NE scores $0$ on all seeds of this task and was not timed; the \napo$^\dagger$ entry is the single timing seed. \quad
\textsuperscript{b}\,Single timed V100 run of AHAC reached $730.0$, whereas AHAC's 3-seed canonical mean is $3.4\pm 4.8$.
\end{table*}

\paragraph{Per-task and fixed configurations in \cref{tab:dmc}.}
Per-task \napo selects from the 8-arm estimator$\times\sigma$ grid in \cref{tab:supp-dmc-grid}, plus $H\in\{16,32\}$, $B=1024$ on the 3 long-contact or sparse-reward tasks. PPO uses MuJoCo Playground's official per-task hyperparameters in \engine and Brax; FastTD3 uses its authors' MuJoCo Playground fork with default hyperparameters (15--41M environment steps) and the same 10-episode evaluation; SHAC~\cite{xu2022shac} and AHAC~\cite{georgiev2024ahac} are ported from \texttt{NVlabs/DiffRL} and \texttt{imgeorgiev/DiffRL} with their published settings ($B=64$, $C=500$). The fixed configuration \napo$^\dagger$ is listed in \cref{tab:supp-dmc-configs}. AHAC is not run on \texttt{Brax ant}. Per-task \napo reaches $0.8\,R^*$ before tuned PPO on 14 of the 16 tasks and has the highest mean final return on 13 of 16; \napo$^\dagger$ has a higher final return than PPO~NE on 10 of 16 tasks and reaches $0.8\,R^*$ first on 12 of the 13 timed tasks.

\section{Additional Humanoid Skills and Analytic Surrogates}
\label{supp:skills}

\begin{table*}[t]
\centering
\scriptsize
\caption{\textbf{Controlled comparisons of analytic surrogate gradients on 29-DoF Unitree G1 humanoid--object interaction (HOI: box pushing across randomized box sizes and 2\,kg box lifting) and 10-type terrain traversal.} Arms share the forward simulator (\engine) and evaluation and differ only in the backward VJP or training objective.}
\label{tab:supp-hoi}
\setlength{\tabcolsep}{4.5pt}
\resizebox{\textwidth}{!}{%
\begin{tabular}{@{}lllcccc@{}}
\toprule
Task & Gradient / objective arm & Forward pass & Backward pass & Seeds & Primary metric & Position / orientation error \\
\midrule
HOI: Box push & Exact objective + exact adjoint & Exact \engine & Exact adjoint & 6 & $68\pm 14\%$ success & $6.5\pm 0.9$\,cm mean ($23.9$\,cm max) \\
(29-DoF G1) & Smoothed contact objective (annealed) & Exact \engine & Exact of soft loss & 6 & $\mathbf{89\pm 2\%}$ success & $\mathbf{2.9\pm 0.2}$\,\textbf{cm} ($9.3$\,cm max, $p=0.012$) \\
\midrule
HOI: Box push & Soft-contact loss + exact backward & Exact \engine & Exact adjoint & 3 & $79\%$ in-dist.\ ($64\%$ OOD) & $4.6$\,cm mean box pos.\ error \\
(0.4--1.6$\times$ size) & \textbf{+ Pre-contact impulse VJP (\cref{eq:surrogate})} & \textbf{Bitwise exact} & \textbf{Impulse VJP} & 3 & $\mathbf{83\%}$ in-dist.\ ($\mathbf{71\%}$ OOD) & $\mathbf{3.8}$\,\textbf{cm} ($p\approx 0.03$) \\
\midrule
HOI: 2\,kg box lift & Exact adjoint (no contact smoothing) & Exact \engine & Exact adjoint & 4 & $\mathbf{99\pm 1\%}$ (in $8.8$--$10.2$\,min) & $2.6$\,cm pos., $4.5^\circ$ mean ($14.3^\circ$ max) tilt \\
(OMOMO clip) & Smoothed contact objective & Exact \engine & Exact of soft loss & 4 & $\mathbf{99\pm 1\%}$ (in $8.9$--$9.2$\,min) & $\mathbf{2.2}$\,\textbf{cm} pos., $\mathbf{3.3^\circ}$ mean ($\mathbf{6.5^\circ}$ max, $p<10^{-3}$) \\
\midrule
Terrain (10$\times$10) & Unsmoothed height-scan VJP ($\partial z/\partial xy = 0$ on flats) & Exact \engine & Zero on flat cells & 320 poses & Cosine $0.26$--$0.36$ vs.\ ref. & Blind to step / platform / gap edges \\
 & \textbf{Gaussian-smoothed height-scan VJP ($\sigma=1$ cell)} & \textbf{Bitwise exact} & \textbf{Smoothed VJP} & 320 poses & \textbf{Cosine $0.985$} ($1.000$ slopes) & Exposes step edges \\
\bottomrule
\end{tabular}}%
\end{table*}

\paragraph{Specifications of the four additional Unitree G1 skills (\cref{tab:tasks,fig:tasks}) and analytic surrogates (\cref{tab:supp-hoi}).}
\begin{enumerate}[leftmargin=*,itemsep=1pt,topsep=2pt]
  \item \textbf{Unified joystick control and LAFAN1 motion tracking:} A single policy trained from scratch without distillation on one A100 GPU (1 seed; $B=4096, H=64, 0.894$\,s/update) that accepts a mode bit to either walk on omnidirectional velocity commands ($v_x \in [-0.8, 1.2]$\,m/s, $v_y \in [-0.5, 0.5]$\,m/s, $\omega_z \in [-1.2, 1.2]$\,rad/s) or track upright LAFAN1 motion clips from standing. It meets the acceptance criterion (0 falls on a 33-command standing grid with median linear velocity error $\le 5$\,cm/s (measured $1.7$\,cm/s), $\le 5\%$ falls on 20\,s LAFAN1 windows (measured $2.1\%$), and $0/32$ falls under pushes) at update 750 ($11.3$\,min), and completes the full velocity sweep without falls at $112.7$\,min ($1.1$\,cm/s velocity error, $2.9$\,cm body position error).
  \item \textbf{Omnidirectional velocity tracking:} Trained on one RTX PRO 6000 GPU per seed (3 seeds, $B=2048, H=64, 0.32$\,s/update) with domain randomization ($K_p \times \mathcal{U}[0.8, 1.2]$, 1-step action delay on $50\%$ of episodes, external pushes). All 3 seeds satisfy the success criterion (0 falls, median linear velocity error $\le 0.05$\,m/s, yaw error $\le 0.10$\,rad/s, foot slip $\le 0.03$\,m/s up to a $1.43$\,m/s fast walk) at the first evaluated checkpoint (update 500, $2.7$--$2.8$\,min), converging to $0.006$\,m/s linear and $0.030$\,rad/s yaw error.
  \item \textbf{Multi-terrain traversal (10 types $\times$ 10 levels):} Trained on one A40 GPU per seed (3 seeds, $B=2048, 0.6$\,s/update) over 10 terrain types (flat, rough, slopes up to $70\%$, stairs up/down with $23$\,cm risers, $45$\,cm platforms and pits, boxes, stepping blocks, and $70$\,cm gaps) from proprioception and a $16\times 9$ ($0.1$\,m) height scan. The 3 seeds cross $\ge 99\%$ of all 100 terrain cells in $18.7$, $18.8$, and $22.4$\,min, reaching $99.8\%$ across all 3 seeds at 40\,min and $98.4\%$--$99.4\%$ on held-out layouts.
  \item \textbf{Humanoid--object interaction (HOI: 2\,kg box lift \& box push, \cref{tab:supp-hoi}):} On 29-DoF G1 box lifting (tracking a 13\,s OMOMO human--box trajectory on one H200 GPU, 4 seeds, $B=4096, H=32, 0.71$--$0.76$\,s/update), all 4 seeds reach $\ge 90\%$ strict success (no fall, box position error $<0.2$\,m, orientation error $<30^\circ$ at every step) in $8.8$, $8.9$, $9.4$, and $10.2$\,min, converging to $99\pm 1\%$ success with $2.6$\,cm mean box position error. \Cref{tab:supp-hoi} summarizes the controlled comparisons between exact adjoints, the straight-through pre-contact impulse surrogate (\cref{eq:surrogate}, akin to straight-through contacts-from-distance~\cite{paulus2026diffmjx}), smoothed contact objectives, and the Gaussian-smoothed height-scan VJP.
\end{enumerate}

\FloatBarrier

\end{document}